\documentclass{article}
\usepackage{iclr2026_conference,times}

\usepackage{amsmath,amsfonts,bm}

\def\eqref#1{equation~\ref{#1}}

\def\1{\bm{1}}

\DeclareMathAlphabet{\mathsfit}{\encodingdefault}{\sfdefault}{m}{sl}
\SetMathAlphabet{\mathsfit}{bold}{\encodingdefault}{\sfdefault}{bx}{n}

\usepackage{hyperref}
\usepackage{url}
\usepackage{times}  
\usepackage{graphicx}
\usepackage{xspace}
\usepackage{balance}
\usepackage{flushend}
\usepackage{soul}
\usepackage{mwe}
\usepackage{subcaption}
\usepackage{multirow}

\usepackage{booktabs}
\usepackage{colortbl}
\usepackage{xcolor}

\usepackage{amssymb}
\usepackage{pifont}

\usepackage{tcolorbox}
\usepackage{enumitem}
\usepackage{wrapfig}

\usepackage{fontawesome5}
\usepackage{placeins}

\tcbuselibrary{listings}
\usepackage{listings}

\hypersetup{
  colorlinks=true,
  linkcolor=black,
  urlcolor=black,
  citecolor=purple
}

\title{Agentic Context Engineering: Evolving Contexts for Self-Improving Language Models}

\author{Qizheng Zhang$^{1*}$, Changran Hu$^{2*}$, Shubhangi Upasani$^2$, Boyuan Ma$^2$, Fenglu Hong$^2$, \\
\textbf{Vamsidhar Kamanuru$^2$, Jay Rainton$^2$, Chen Wu$^2$, Mengmeng Ji$^2$, Hanchen Li$^3$,} \\
\textbf{Urmish Thakker$^2$, James Zou$^1$, Kunle Olukotun$^1$} \\
[0.2em]
$^1$Stanford University \quad $^2$SambaNova Systems, Inc. \quad $^3$UC Berkeley\\
[0.5em]
{\ttfamily\small
\{\href{mailto:qizhengz@stanford.edu}{qizhengz}%
,\href{mailto:kunle@stanford.edu}{kunle}%
\}@stanford.edu
\quad
\href{mailto:changran_hu@berkeley.edu}{\nolinkurl{changran_hu@berkeley.edu}}
}\\
[0.2em]
{\ttfamily\small
\raisebox{-0.1em}{\faGithub}\ \href{https://github.com/ace-agent/ace}{\texttt{ace-agent/ace}} \quad
\raisebox{-0.1em}{\faGlobe}\ \href{https://ace-agent.github.io}{\texttt{ace-agent.github.io}}
}\quad
{\footnotesize $^{*}$Equal contribution.}
}



\iclrfinalcopy
\begin{document}

\maketitle


\newcommand{\name}{\textsc{ACE}\xspace}

\newcommand{\fillme}{{\bf XXX}\xspace}
\newcommand{\eg}{{\it e.g.,}\xspace}
\newcommand{\ie}{{\it i.e.,}\xspace}
\newcommand{\etal}{{\it et.~al}\xspace}
\newcommand{\bigO}{\mathrm{O}}

\newcommand{\cmark}{\ding{51}}  
\newcommand{\xmark}{\ding{55}}  

\newcounter{packednmbr}
\newenvironment{packedenumerate}{\begin{list}{\thepackednmbr.}{\usecounter{packednmbr}\setlength{\itemsep}{0.5pt}\addtolength{\labelwidth}{-4pt}\setlength{\leftmargin}{2ex}\setlength{\listparindent}{\parindent}\setlength{\parsep}{1pt}\setlength{\topsep}{2pt}}}{\end{list}}
\newenvironment{packeditemize}{\begin{list}{$\bullet$}{\setlength{\itemsep}{0.5pt}\addtolength{\labelwidth}{-4pt}\setlength{\leftmargin}{2ex}\setlength{\listparindent}{\parindent}\setlength{\parsep}{1pt}\setlength{\topsep}{2pt}}}{\end{list}}
\newenvironment{packedpackeditemize}{\begin{list}{$\bullet$}{\setlength{\itemsep}{0.5pt}\addtolength{\labelwidth}{-4pt}\setlength{\leftmargin}{\labelwidth}\setlength{\listparindent}{\parindent}\setlength{\parsep}{1pt}\setlength{\topsep}{0pt}}}{\end{list}}
\newenvironment{packedtrivlist}{\begin{list}{\setlength{\itemsep}{0.2pt}\addtolength{\labelwidth}{-4pt}\setlength{\leftmargin}{\labelwidth}\setlength{\listparindent}{\parindent}\setlength{\parsep}{1pt}\setlength{\topsep}{0pt}}}{\end{list}}

\definecolor{refinegreen}{RGB}{0, 128, 75}
\definecolor{scoregreen}{RGB}{34, 139, 34}

\definecolor{darkgray}{RGB}{70,70,70}
\definecolor{lightgray}{RGB}{240,240,240}

\tcbset{
    boxstyle/.style={
        enhanced,
        sharp corners,
        colback=gray!20,
        colframe=gray!60,
        boxrule=1pt,
        left=10pt,
        right=10pt,
        top=5pt,
        bottom=5pt
    },
    headerstyle/.style={
        enhanced,
        sharp corners,
        colback=gray!60,
        coltext=white,
        boxrule=0pt,
        left=10pt,
        right=10pt,
        top=5pt,
        bottom=5pt,
        fontupper=\bfseries
    },
    verbatimstyle/.style={
        enhanced,
        sharp corners,
        colback=gray!10,
        colframe=gray!40,
        boxrule=1pt,
        left=10pt,
        right=10pt,
        top=5pt,
        bottom=5pt,
        listing only,
        fontupper=\ttfamily\small
    }
}

\definecolor{PromptBack}{gray}{0.97}
\definecolor{PromptTitleBack}{gray}{0.90}
\definecolor{PromptFrame}{gray}{0.60}

\lstdefinelanguage{prompt}{
  morekeywords={USER,SYSTEM,ASSISTANT,Task,Context,Cheatsheet,Reflection,Instructions,Answer},
  sensitive=false
}

\newcommand{\lstb}[1]{\textbf{#1}}
\newcommand{\lstem}[1]{\textit{#1}}

\lstdefinestyle{promptstyle}{
  language=prompt,
  basicstyle=\ttfamily\footnotesize,
  numbers=left, numberstyle=\tiny, numbersep=9pt, stepnumber=1,
  columns=fullflexible, keepspaces=true, showstringspaces=false,
  breaklines=true, breakatwhitespace=true,
  frame=none,
  escapeinside={(*@}{@*)},
  literate=
    {…}{{\ldots}}1
    {–}{{--}}1
    {—}{{---}}1
    {“}{{``}}1
    {”}{{''}}1
    {’}{{'}}1
    {•}{{\textbullet}}1
}

\newtcblisting{PromptCard}[2][]{%
  listing engine=listings, listing only,
  colback=PromptBack, colframe=PromptFrame, boxrule=0.5pt,
  arc=1.2mm,
  left=2mm,right=2mm,top=1.2mm,bottom=1.2mm,
  listing options={style=promptstyle, xleftmargin=1.8em}, 
  title={#2}, fonttitle=\bfseries\sffamily\footnotesize,
  colbacktitle=PromptTitleBack, coltitle=black,
  #1
}

\newcommand{\PromptCardFromFile}[3][]{%
  \begin{tcolorbox}[
    colback=PromptBack, colframe=PromptFrame, boxrule=0.5pt,
    arc=1.2mm, left=2mm,right=2mm,top=1.2mm,bottom=1.2mm,
    title={#2}, fonttitle=\bfseries\sffamily\footnotesize,
    colbacktitle=PromptTitleBack, coltitle=black,
    #1
  ]
    \lstinputlisting[style=promptstyle, xleftmargin=1.8em]{#3}
  \end{tcolorbox}
}

\begin{abstract}
Large language model (LLM) applications such as agents and domain-specific reasoning increasingly rely on \emph{context adaptation}: modifying inputs with instructions, strategies, or evidence, rather than weight updates. 
Prior approaches improve usability but often suffer from brevity bias, which drops domain insights for concise summaries, and from context collapse, where iterative rewriting erodes details over time. 
We introduce ACE (\textbf{A}gentic \textbf{C}ontext \textbf{E}ngineering), a framework that treats contexts as evolving playbooks that accumulate, refine, and organize strategies through a modular process of generation, reflection, and curation. 
ACE prevents collapse with structured, incremental updates that preserve detailed knowledge and scale with long-context models.
Across agent and domain-specific benchmarks, ACE optimizes contexts both offline (\eg system prompts) and online (\eg agent memory), consistently outperforming strong baselines: +10.6\% on agents and +8.6\% on finance, while significantly reducing adaptation latency and rollout cost. 
Notably, ACE could adapt effectively without labeled supervision and instead by leveraging natural execution feedback. 
On the AppWorld leaderboard, ACE matches the top-ranked production-level agent on the overall average and surpasses it on the harder test-challenge split, despite using a smaller open-source model.
These results show that comprehensive, evolving contexts enable scalable, efficient, and self-improving LLM systems with low overhead.
\end{abstract}
\section{Introduction}

\begin{figure}[htbp]
    \centering
    \includegraphics[width=0.95\textwidth]{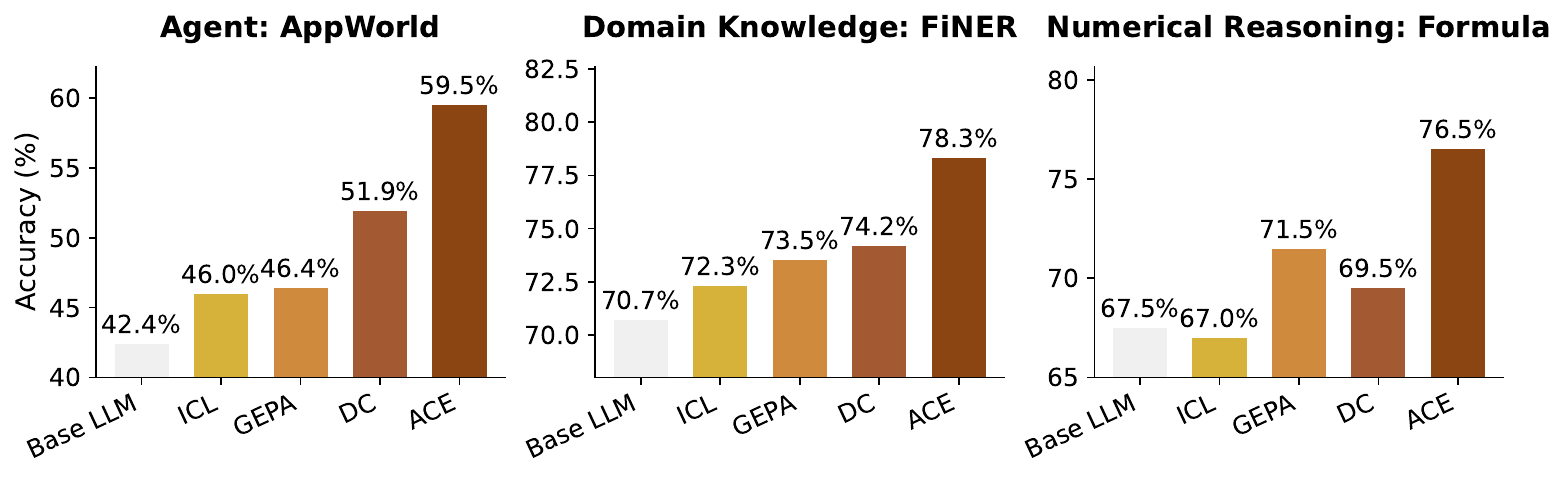}
    \caption{\textbf{Overall Performance Results.} Our proposed framework, ACE, consistently outperforms strong baselines across agent and domain-specific tasks.}
    \label{fig:overall-gain}
\end{figure}

Modern AI applications based on large language models (LLMs), such as LLM agents~\citep{yao2023react, yang2024swe} and compound AI systems~\citep{zaharia2024compoundGS}, increasingly depend on \textit{context adaptation}.
Instead of modifying model weights, context adaptation improves performance after model training by incorporating clarified instructions, structured reasoning steps, or domain-specific input formats directly into the model's inputs.
Contexts underpin many AI system components, including system prompts that guide downstream tasks~\citep{opsahl2024optimizing, agrawal2025gepa}, memory that carries past facts and experiences~\citep{suzgun2025dynamic, xu2025mem}, and factual evidence that reduces hallucination and supplements knowledge~\citep{asai2024self}.

Adapting through \textit{contexts} rather than \textit{weights} offers several key advantages.
Contexts are interpretable and explainable for users and developers~\citep{wei2022chain, wang2022self}, allow rapid integration of new knowledge at runtime~\citep{lewis2020retrieval, borgeaud2022improving}, and can be shared across models or modules in a compound system~\citep{khot2022decomposed}.
Meanwhile, advances in long-context LLMs~\citep{peng2023yarn} and context-efficient inference such as KV cache reuse~\citep{gim2024prompt, yao2025cacheblend} are making context-based approaches increasingly practical for deployment.
As a result, context adaptation is emerging as a central paradigm for building capable, scalable, and self-improving AI systems.

\begin{figure}[htbp]
    \centering
    \includegraphics[width=0.8\textwidth]{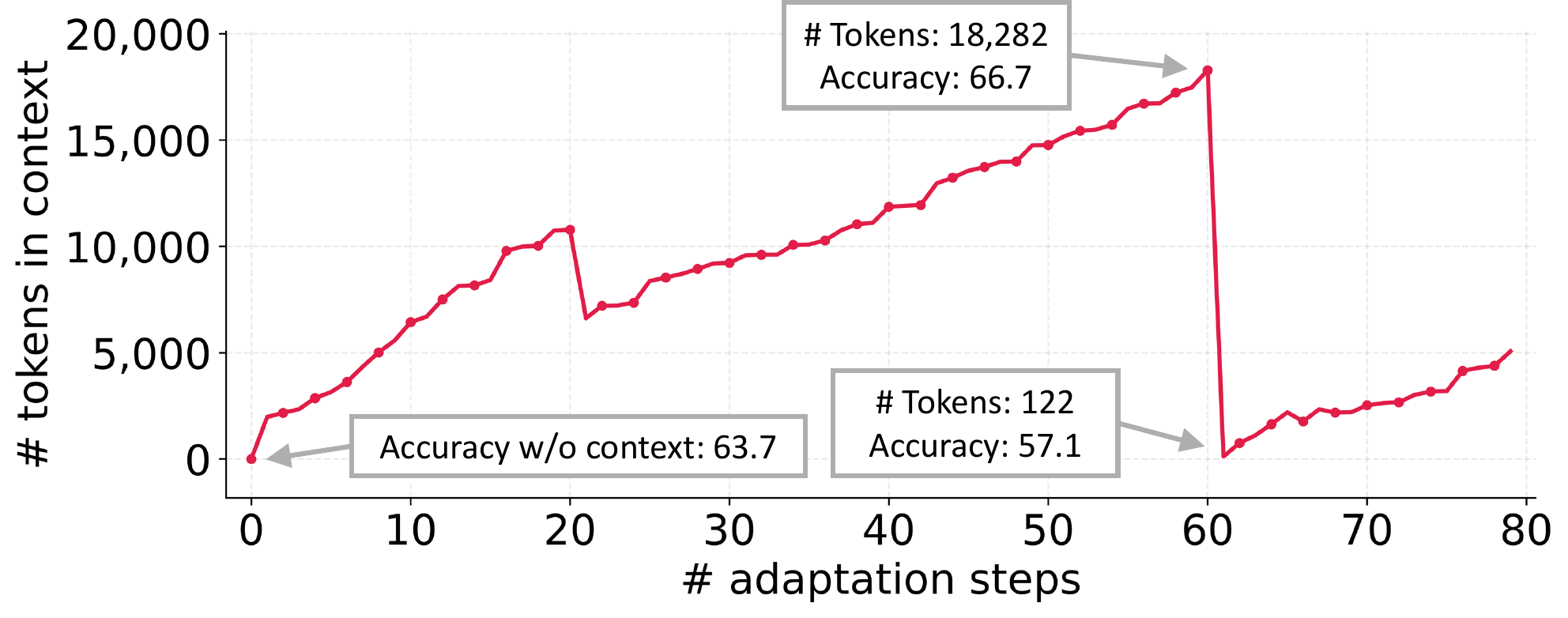}
    \caption{\textbf{Context Collapse.} Monolithic rewriting of context by an LLM can collapse it into shorter, less informative summaries, leading to sharp performance drops. }
    \label{fig:context-collapse}
\end{figure}

Despite this progress, existing approaches to context adaptation face two limitations.
First, \textit{brevity bias}: many prompt optimizers prioritize concise applicable instructions over comprehensive accumulation.
For example, GEPA~\citep{agrawal2025gepa} highlights brevity as a strength, but such abstraction can omit domain-specific heuristics, tool-use guidelines, or common failure modes that matter in practice~\citep{gao2025prompt}.
This objective aligns with validation metrics in some settings, but often fails to capture the detailed strategies required by agents and knowledge-intensive applications.
Second, \textit{context collapse}: methods that rely on monolithic rewriting by an LLM often degrade into shorter, less informative summaries over time, causing sharp performance declines (Figure \ref{fig:context-collapse}).
In domains such as interactive agents~\citep{trivedi2024appworld, patil2024gorilla, zhang2024caravan}, domain-specific programming~\citep{ye2023generating, zhang2025adaptive, zhang2025accelopt, mang2025frontiercs}, and financial or legal analysis~\citep{loukas2022finer, guha2023legalbench, wang2025finlora}, strong performance depends on retaining detailed, task-specific knowledge rather than compressing it away.

As applications like agents and knowledge-intensive reasoning demand greater reliability, recent work has shifted toward saturating contexts with abundant, potentially useful information~\citep{jiang2025putting, chung2025long, chen2025flora}, enabled by advances in long-context LLMs~\citep{peng2023yarn, mao2024lift}.
\textbf{We argue that contexts should function not as concise summaries, but as comprehensive, structured playbooks that are detailed, inclusive, and rich with domain insights.}
Unlike humans, who often benefit from concise generalization, LLMs are more effective when provided with long, detailed contexts and can distill relevance autonomously~\citep{jiang2025putting, liu2025selfelicit, suzgun2025dynamic}.
Thus, instead of compressing away domain-specific heuristics and tactics, contexts should preserve them, allowing the model to decide what matters during inference time.

To address these limitations, we introduce \name (\textbf{A}gentic \textbf{C}ontext \textbf{E}ngineering), a framework for comprehensive context adaptation in both offline settings (\eg system prompt optimization) and online settings (\eg test-time memory adaptation).
Rather than compressing contexts into distilled summaries, \name treats them as evolving playbooks that accumulate and organize strategies over time.
By design, \name incorporates a modular workflow of generation, reflection, and curation, while adding structured, incremental updates guided by a grow-and-refine principle.
This design preserves detailed, domain-specific knowledge, prevents context collapse, and yields contexts that remain comprehensive and scalable throughout adaptation.

We evaluate \name on two categories of LLM applications that most benefit from comprehensive, evolving contexts:
(1) \emph{agents}~\citep{trivedi2024appworld}, which require multi-turn reasoning, tool use, and environment interaction, where accumulated strategies can be reused across episodes; and
(2) \emph{domain-specific benchmarks}, which demand specialized tactics and knowledge, like financial analysis~\citep{loukas2022finer, wang2025finlora}.
Our key findings are:
\begin{packeditemize}
    \item \name consistently outperforms strong baselines, yielding average gains of 10.6\% on \textit{agents} and 8.6\% on \textit{domain-specific benchmarks}, across both offline and online adaptation settings.
    \item \name is able to construct effective contexts \textit{without} labeled supervision, instead leveraging execution feedback and environment signals, key ingredients for self-improving LLMs and agents.
    \item On the AppWorld benchmark leaderboard~\citep{AppWorldLeaderboard}, \name surpasses the top-1-ranked production-level agent IBM-CUGA~\citep{marreed2025towards} (powered by GPT-4.1) while using an open-source model (DeepSeek-V3.1).
    \item \name requires significantly fewer rollouts and achieves lower adaptation latency than existing adaptive methods, demonstrating that scalable self-improvement can be achieved with both higher accuracy and lower cost.
\end{packeditemize}
\section{Background and Motivation}

\subsection{Context Adaptation}

Context adaptation (or context engineering) refers to methods that improve model behavior by constructing or modifying inputs to an LLM, rather than altering its weights.
The current state of the art leverages \textit{natural language feedback}~\citep{shinn2023reflexion, yuksekgonul2024textgrad, agrawal2025gepa}.
In this paradigm, a language model inspects the current context along with signals such as execution traces, reasoning steps, or validation results, and generates natural language feedback on how the context should be revised.
This feedback is then incorporated into the context, enabling iterative adaptation.
Representative methods include Reflexion~\citep{shinn2023reflexion}, which reflects on failures to improve agent planning; TextGrad~\citep{yuksekgonul2024textgrad}, which optimizes prompts via gradient-like textual feedback; 
GEPA~\citep{agrawal2025gepa}, which refines prompts iteratively based on execution traces and achieves strong performance, even surpassing reinforcement learning approaches in some settings;
and Dynamic Cheatsheet~\citep{krause2019dynamic}, which constructs an external memory that accumulates strategies and lessons from past successes and failures during inference.
These natural language feedback methods represent a major advance, offering flexible and interpretable signals for improving LLM systems beyond weight updates.

\subsection{Limitations of Existing Context Adaptation Methods}
\label{subsec:limitations}

\paragraph{Brevity Bias}
A recurring limitation of context adaptation methods is \textit{brevity bias}: the tendency of optimization to collapse toward short, generic prompts.
Gao et al.~\citep{gao2025prompt} document this effect in prompt optimization for test generation, where iterative methods repeatedly produced near-identical instructions (\eg, “Create unit tests to ensure methods behave as expected”), sacrificing diversity and omitting domain-specific detail.
This convergence not only narrows the search space but also propagates recurring errors across iterations, since optimized prompts often inherit the same faults as their seeds.
More broadly, such bias undermines performance in domains that demand detailed, context-rich guidance—such as multi-step agents, program synthesis, or knowledge-intensive reasoning—where success hinges on accumulating rather than compressing task-specific insights.

\paragraph{Context Collapse} 
In a case study on the AppWorld benchmark~\citep{trivedi2024appworld}, we observe a phenomenon we call \textit{context collapse}, which arises when an LLM is tasked with fully rewriting the accumulated context at each adaptation step. 
As the context grows large, the model tends to compress it into much shorter, less informative summaries, causing a dramatic loss of information. 
For instance, at step 60 the context contained 18,282 tokens and achieved an accuracy of 66.7, but at the very next step it collapsed to just 122 tokens, with accuracy dropping to 57.1—worse than the baseline accuracy of 63.7 without adaptation. 
While we highlight this through Dynamic Cheatsheet~\citep{suzgun2025dynamic}, the issue is not specific to that method; rather, it reflects a fundamental risk of end-to-end context rewriting with LLMs, where accumulated knowledge can be abruptly erased instead of preserved.

\begin{figure}[htbp]
    \centering
    \includegraphics[width=0.9\textwidth]{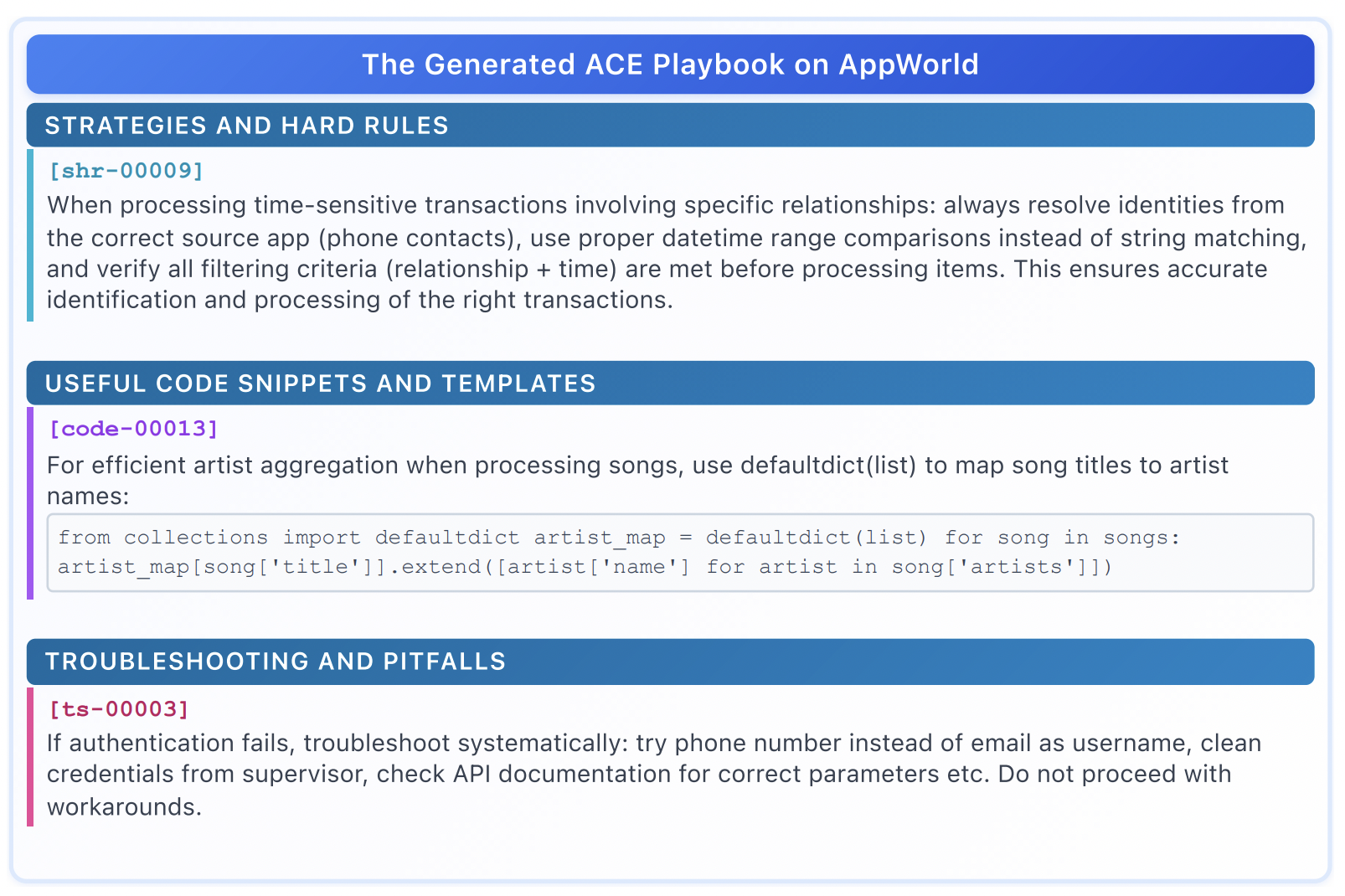}
    \caption{\textbf{Example ACE-Generated Context on the AppWorld Benchmark} (partially shown). ACE-generated contexts contain detailed, domain-specific insights along with tools and code that are readily usable, serving as a comprehensive playbook for LLM applications.}
\label{fig:ace-context-appworld-motivation}
\end{figure}
\section{Agentic Context Engineering (ACE)}
\label{sec:methods}

\begin{figure}[!htbp]
    \centering
    \includegraphics[width=0.95\textwidth]{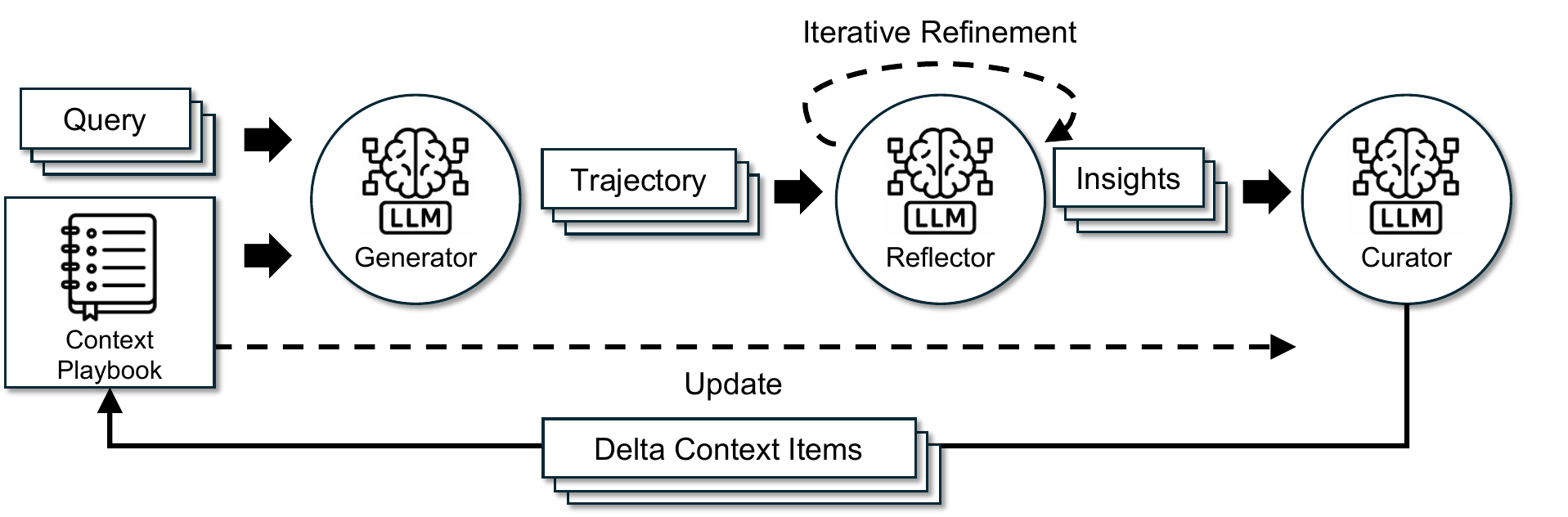}
    \caption{\textbf{The \name Framework.} Inspired by Dynamic Cheatsheet, ACE adopts an agentic architecture with three specialized components: a Generator, a Reflector, and a Curator.}
    \label{fig:design}
\end{figure}

We present ACE (\textbf{A}gentic \textbf{C}ontext \textbf{E}ngineering), a framework for scalable and efficient context adaptation in both offline (\eg system prompt optimization) and online (\eg test-time memory adaptation) scenarios. 
Instead of condensing knowledge into terse summaries or static instructions, ACE treats contexts as evolving playbooks that continuously accumulate, refine, and organize strategies over time. 
Inspired by the agentic design of Dynamic Cheatsheet~\citep{suzgun2025dynamic}, ACE introduces a structured division of labor across three roles (Figure \ref{fig:design}): the \emph{Generator}, which produces reasoning trajectories; the \emph{Reflector}, which distills concrete insights from successes and errors; and the \emph{Curator}, which integrates these insights into structured context updates. 
This mirrors how humans learn: experimenting, reflecting, and consolidating, while avoiding the bottleneck of overloading a single model with all responsibilities.  

To address the limitations of prior methods discussed in \S\ref{subsec:limitations} (notably \textit{brevity bias} and \textit{context collapse}) ACE introduces three key innovations:  
(1) a dedicated \textit{Reflector} that separates evaluation and insight extraction from curation, improving context quality and downstream performance (\S\ref{subsec:results-ablation});  
(2) incremental \emph{delta updates} (\S\ref{subsec:delta-update}) that replace costly monolithic rewrites with localized edits, reducing both latency and compute cost (\S\ref{subsec:results-cost-speed}); and  
(3) a \emph{grow-and-refine} mechanism (\S\ref{subsec:grow-and-refine}) that balances steady context expansion with redundancy control.  

As shown in Figure~\ref{fig:design}, the workflow begins with the Generator producing reasoning trajectories for new queries, which surface both effective strategies and recurring pitfalls. 
The Reflector critiques these traces to extract lessons, optionally refining them across multiple iterations. 
The Curator then synthesizes these lessons into compact \emph{delta entries}, which are merged deterministically into the existing context by lightweight, non-LLM logic. 
Because updates are itemized and localized, multiple deltas can be merged in parallel, enabling batched adaptation at scale. 
ACE further supports multi-epoch adaptation, where the same queries are revisited to progressively strengthen the context.  

\subsection{Incremental Delta Updates}
\label{subsec:delta-update}

A core design principle of ACE is to represent context as a collection of \emph{structured, itemized bullets}, rather than a single monolithic prompt. 
The concept of a bullet is similar to the concept of a memory entry in LLM memory frameworks like Dynamic Cheatsheet~\citep{suzgun2025dynamic} and A-MEM~\citep{xu2025mem}, but builds on top of that and consists of (1) \textbf{metadata}, including a unique identifier and counters tracking how often it was marked helpful or harmful; and (2) \textbf{content}, capturing a small unit such as a reusable strategy, domain concept, or common failure mode. 
When solving new problems, the Generator highlights which bullets were useful or misleading, providing feedback that guides the Reflector in proposing corrective updates. 

This itemized design enables three properties: 
(1) \emph{localization}, so only the relevant bullets are updated; 
(2) \emph{fine-grained retrieval}, so the Generator can focus on the most pertinent knowledge; and 
(3) \emph{incremental adaptation}, allowing efficient merging, pruning, and de-duplication during inference.

Rather than regenerating contexts in full, ACE incrementally produces compact \emph{delta contexts}: small sets of candidate bullets distilled by the Reflector and integrated by the Curator. 
This avoids the computational cost and latency of full rewrites, while ensuring that past knowledge is preserved and new insights are steadily appended. 
As contexts grow, this approach provides the scalability needed for long-horizon or domain-intensive applications.

\subsection{Grow-and-Refine}
\label{subsec:grow-and-refine}

Beyond incremental growth, ACE ensures that contexts remain compact and relevant through periodic or lazy refinement. 
In grow-and-refine, bullets with new identifiers are appended, while existing bullets are updated in place (\eg incrementing counters). 
A de-duplication step then prunes redundancy by comparing bullets via semantic embeddings. 
This refinement can be performed proactively (after each delta) or lazily (only when the context window is exceeded), depending on application requirements for latency and accuracy.

Together, incremental updates and grow-and-refine maintain contexts that expand adaptively, remain interpretable, and avoid the potential variance introduced by monolithic context rewriting.
\section{Results}

Our evaluation of \name shows that:
\begin{packeditemize}
    \item \textbf{Enabling High-Performance, Self-Improving Agents.} ACE enables agents to self-improve by dynamically refining their input context, both in offline and online settings. It boosts accuracy on the AppWorld benchmark by up to 17.1\% by learning to engineer better contexts from execution feedback alone, without needing ground-truth labels. (\S\ref{subsec:agent-main-results})
    \item \textbf{Large Gains on Domain-Specific Benchmarks.} On complex financial reasoning benchmarks, ACE delivers an average performance gain of 8.6\% over strong baselines by constructing comprehensive playbooks with domain-specific concepts and insights. (\S\ref{subsec:finance-main-results})
    \item \textbf{Effective by Design.} Ablation studies confirm our design choices are key to success, with components like the Reflector, multi-epoch refinement, and incremental delta update each contributing substantial performance gains. (\S\ref{subsec:results-ablation})
    \item \textbf{Lower Cost and Adaptation Latency.} ACE achieves these gains efficiently, reducing adaptation latency by 86.9\% on average, while requiring fewer rollouts and lower token dollar costs. (\S\ref{subsec:results-cost-speed})
\end{packeditemize}

\subsection{Tasks and Datasets}

We evaluate \name on two categories of LLM applications that benefit most from evolving contexts: 
(1) \emph{LLM agent}, which require multi-turn reasoning, tool use, and environment interaction; with ACE, agents can accumulate and reuse strategies across episodes and environments; and  
(2) \emph{domain-specific reasoning}, which demand mastery of specialized concepts and tactics; we focus on financial analysis as a main case study, and show additional results on medical reasoning and text-to-SQL. 

\begin{packeditemize}
    \item \textbf{LLM Agent: AppWorld~\citep{trivedi2024appworld}} is a suite of autonomous agent tasks involving API understanding, code generation, and environment interaction. 
    It provides a realistic execution environment with common applications and APIs (\eg email, file system) and tasks of two difficulty levels (normal and challenge). 
    A public leaderboard~\citep{AppWorldLeaderboard} tracks performance, where, at the time of submission, the best system achieved only 60.3\% average accuracy, highlighting the benchmark’s difficulty and realism.
    \item \textbf{Domain-Specific Reasoning: Financial, Medical, and Text-to-SQL Benchmarks} We use finance as our main case study in \S\ref{subsec:finance-main-results}. 
    For financial analysis, we focus on FiNER~\citep{loukas2022finer} and Formula~\citep{wang2025finlora}, which test LLMs on financial reasoning tasks that rely on the eXtensible Business Reporting Language (XBRL).
    \emph{FiNER} requires labeling tokens in XBRL financial documents with one of 139 fine-grained entity types, a key step for financial information extraction in regulated domains.  
    \emph{Formula} focuses on applying financial concepts and performing computations to answer queries, \ie numerical reasoning.
    Beyond finance, we evaluate on two additional domain tasks from StreamBench~\citep{wu2024streambench}: DDXPlus~\citep{fansi2022ddxplus} (medical reasoning) and BIRD-SQL~\citep{li2023can} (text-to-SQL).
\end{packeditemize} 

\paragraph{Evaluation Metrics}  
For AppWorld, we follow the official benchmark protocol and report \emph{Task Goal Completion} (TGC) and \emph{Scenario Goal Completion} (SGC) on both the test-normal and test-challenge splits.  
For FiNER, Formula and DDXPlus, we follow the original setup and report accuracy, measured as the proportion of predicted answers that exactly match the ground truth. 
For BIRD-SQL, we use GPT-4o-mini~\citep{gpt4omini} under LLM-as-a-judge~\citep{zheng2023judging}. 

All datasets follow the original train/validation/test splits. 
For \emph{offline} context adaptation, methods are optimized on the training split and evaluated on the test split with pass@1 accuracy.  
For \emph{online} context adaptation, methods are evaluated sequentially on the test split: for each sample, the model first predicts with the current context, then updates its context based on that sample.  
The same shuffled test split is used across all methods.

\subsection{Baselines and Methods}

\paragraph{Base LLM} The base model is evaluated directly on each benchmark without any context engineering, using the default prompts provided by dataset authors.  
For AppWorld, we follow the official \texttt{ReAct}~\citep{yao2023react} implementation released by the benchmark authors, and build all other baselines and methods on top of this framework. 

\paragraph{In-Context Learning (ICL)~\citep{agarwal2024many}} ICL provides the model with task demonstrations in the input prompt (few-shot or many-shot). 
This allows the model to infer the task format and desired output without weight updates.  
We supply all training samples when they fit within the model’s context window; otherwise, we fill the window with as many demonstrations as possible.

\paragraph{MIPROv2~\citep{opsahl2024optimizing}} MIPROv2 is a popular prompt optimizer for LLM applications that works by jointly optimizing system instructions and in-context demonstrations via bayesian optimization.   
We use the official DSPy implementation~\citep{DSPyMIPROv2}, setting \verb|auto="heavy"| to maximize optimization performance.

\paragraph{GEPA~\citep{agrawal2025gepa}} GEPA (Genetic-Pareto) is a sample-efficient prompt optimizer based on reflective prompt evolution. 
It collects execution traces (reasoning, tool calls, intermediate outputs) and applies natural-language reflection to diagnose errors, assign credit, and propose prompt updates. 
A genetic Pareto search maintains a frontier of high-performing prompts, mitigating local optima.  
Empirically, GEPA outperforms reinforcement learning methods such as GRPO and prompt optimizers like MIPROv2, achieving up to 10–20\% higher accuracy with as much as 35× fewer rollouts. 
We use the official DSPy implementation~\citep{DSPyGEPA}, setting \verb|auto="heavy"| to maximize optimization performance.

\paragraph{Dynamic Cheatsheet (DC)~\citep{suzgun2025dynamic}} DC is a test-time learning approach that introduces an adaptive external memory of reusable strategies and code snippets.  
By continuously updating this memory with newly encountered inputs and outputs, DC enables models to accumulate knowledge and reuse it across tasks, often leading to substantial improvements over static prompting methods.  
A key advantage of DC is that it does not require ground-truth labels: the model can curate its own memory from its generations, making the method highly flexible and broadly applicable.  
We use the official implementation released by the authors~\citep{Suzgun2025_DynamicCheatsheet_code} and set it to use the \texttt{cumulative} mode (DC-CU).

\paragraph{ACE (ours)} ACE optimizes LLM contexts for both offline and online adaptation through an agentic context engineering framework.  
To ensure fairness, we use the same LLM for the Generator, Reflector, and Curator (non-thinking mode of DeepSeek-V3.1~\citep{deepseekai2024deepseekv3technicalreport}), preventing knowledge transfer from a stronger Reflector or Curator to a weaker Generator. 
This isolates the benefit of context construction itself. 
We additionally evaluate ACE with other backbone LLMs in the appendix, where we observe consistent gains.
We adopt a batch size of 1 (constructing a delta context from each sample).
We set the maximum number of Reflector refinement rounds and the maximum number of epoch in offline adaptation to 5. 

\subsection{Results on Agent Benchmark}
\label{subsec:agent-main-results}

\begin{table*}[!tbhp]
  \centering
  \resizebox{\textwidth}{!}{ 
  \begin{tabular}{lc|cc|cc|c}
    \toprule[1.25pt]
    \multirow{2}{*}{\textbf{Method}} 
      & \multirow{2}{*}{\textbf{GT Labels}}
      & \multicolumn{2}{c|}{\textbf{Test-Normal}} 
      & \multicolumn{2}{c|}{\textbf{Test-Challenge}} 
      & \multirow{2}{*}{\textbf{Average}}\\
    \cmidrule(lr){3-4}\cmidrule(lr){5-6}
      & & \textbf{TGC$\uparrow$} & \textbf{SGC$\uparrow$} & \textbf{TGC$\uparrow$} & \textbf{SGC$\uparrow$} \\
    \midrule[1.1pt]
    \rowcolor[rgb]{0.93,0.93,0.93}
    \multicolumn{7}{c}{\texttt{DeepSeek-V3.1-671B as Base LLM}} \\
    \textcolor{gray}{\texttt{ReAct}} & & \textcolor{gray}{63.7} & \textcolor{gray}{42.9} & \textcolor{gray}{41.5} & \textcolor{gray}{21.6} & \textcolor{gray}{42.4} \\
    \midrule
    \rowcolor[rgb]{0.93,0.93,0.93}
    \multicolumn{7}{c}{\texttt{Offline Adaptation}} \\
    \texttt{ReAct} + ICL & \checkmark & $64.3_{\textcolor{scoregreen}{+0.6}}$ & $46.4_{\textcolor{scoregreen}{+3.5}}$ & $46.0_{\textcolor{scoregreen}{+4.5}}$ & $27.3_{\textcolor{scoregreen}{+5.7}}$ & $46.0_{\textcolor{scoregreen}{+3.6}}$ \\
    \texttt{ReAct} + GEPA & \checkmark & $64.9_{\textcolor{scoregreen}{+1.2}}$ & $44.6_{\textcolor{scoregreen}{+1.7}}$ & $46.0_{\textcolor{scoregreen}{+4.5}}$ & $30.2_{\textcolor{scoregreen}{+8.6}}$ & $46.4_{\textcolor{scoregreen}{+4.0}}$\\
    \texttt{ReAct} + ACE & \checkmark & $\textcolor{black}{\mathbf{76.2}}_{\textcolor{scoregreen}{\mathbf{+12.5}}}$ & $\textcolor{black}{\mathbf{64.3}}_{\textcolor{scoregreen}{\mathbf{+21.4}}}$ & $\textcolor{black}{\mathbf{57.3}}_{\textcolor{scoregreen}{\mathbf{+15.8}}}$ & $\textcolor{black}{\mathbf{39.6}}_{\textcolor{scoregreen}{\mathbf{+18.0}}}$ & $\textcolor{black}{\mathbf{59.4}}_{\textcolor{scoregreen}{\mathbf{+17.0}}}$\\
    \texttt{ReAct} + ACE & \xmark & $75.0_{\textcolor{scoregreen}{+11.3}}$ & $\textcolor{black}{\mathbf{64.3}}_{\textcolor{scoregreen}{\mathbf{+21.4}}}$ & $54.4_{\textcolor{scoregreen}{+12.9}}$ & $35.2_{\textcolor{scoregreen}{+13.6}}$ & $57.2_{\textcolor{scoregreen}{+14.8}}$\\
    \midrule
    \rowcolor[rgb]{0.93,0.93,0.93}
    \multicolumn{7}{c}{\texttt{Online Adaptation}} \\
    \texttt{ReAct} + DC (CU) & \xmark & $65.5_{\textcolor{scoregreen}{+1.8}}$ & $\textcolor{black}{\mathbf{58.9}}_{\textcolor{scoregreen}{\mathbf{+16.0}}}$ & $52.3_{\textcolor{scoregreen}{+10.8}}$ & $30.8_{\textcolor{scoregreen}{+9.2}}$ & $51.9_{\textcolor{scoregreen}{+9.5}}$\\
    \texttt{ReAct} + ACE & \xmark & $\textcolor{black}{\mathbf{69.6}}_{\textcolor{scoregreen}{\mathbf{+5.9}}}$ & $53.6_{\textcolor{scoregreen}{+10.7}}$ & $\textcolor{black}{\mathbf{66.0}}_{\textcolor{scoregreen}{\mathbf{+24.5}}}$ & $\textcolor{black}{\mathbf{48.9}}_{\textcolor{scoregreen}{\mathbf{+27.3}}}$ & $\textcolor{black}{\mathbf{59.5}}_{\textcolor{scoregreen}{\mathbf{+17.1}}}$\\
    \bottomrule[1.25pt]
  \end{tabular}}
  \caption{\textbf{Results on the AppWorld Agent Benchmark (DeepSeek-V3.1-671B as the Base LLM).} ``GT labels" indicates whether ground-truth labels are available to the Reflector during adaptation. We evaluate the ACE framework against multiple baselines on top of the official \texttt{ReAct} implementation, both for offline and online context adaptation. 
  \texttt{ReAct} + ACE outperforms selected baselines by an average of 10.6\%, and could achieve good performance even without access to GT labels. }
  \label{tab:appworld-results}
  \vspace{-5pt}
\end{table*}

\paragraph{Analysis: AppWorld}  
As shown in Table~\ref{tab:appworld-results}, ACE consistently improves over strong baselines on AppWorld.  
In the offline setting, \texttt{ReAct} + ACE outperforms both \texttt{ReAct} + ICL and \texttt{ReAct} + GEPA by significant margins (12.3\% and 11.9\%, respectively), demonstrating that structured, evolving, and detailed contexts enable more effective agent learning than fixed demonstrations or single optimized instruction prompts.  
These gains extend to the online setting, where ACE continues to outperform prior adaptive methods such as Dynamic Cheatsheet by an average of 7.6\%.

In the agent use case, ACE remains effective even \textit{without} access to ground-truth labels during adaptation: \texttt{ReAct} + ACE achieves an average improvement of 14.8\% over the \texttt{ReAct} baseline in this setting.  
This robustness arises because ACE leverages signals naturally available during execution (\eg code execution success or failure) to guide the Reflector and Curator in forming structured lessons of successes and failures. 
Together, these results establish ACE as a strong and versatile framework for building self-improving agents that adapt reliably both with and without labeled supervision. 

Notably, on the latest AppWorld leaderboard (as of September 20, 2025; Figure \ref{fig:appworld-leaderboard}), \texttt{ReAct} + ACE (59.4\% average) matches the top-1-ranked IBM CUGA (60.3\%)\footnote{We mention IBM CUGA as a rough contextual reference to show that ACE operates in a similar performance range on the AppWorld leaderboard. It is not used as a methodological baseline, and we do not make direct comparisons. CUGA’s internal design differs from ACE’s context-adaptation focus, and all baselines are evaluated under identical setups to isolate methodological effects rather than agent-engineering choices.}, a production-level GPT-4.1–based agent~\citep{marreed2025towards}, despite using the much smaller open-source model DeepSeek-V3.1.
With online adaptation, \texttt{ReAct} + ACE even surpasses IBM CUGA by 8.4\% in TGC and 0.7\% in SGC on test-challenge, underscoring the effectiveness of \name in building comprehensive and self-evolving contexts for agents.

\subsection{Results on Domain-Specific Benchmark}
\label{subsec:finance-main-results}

\begin{table*}[!tbhp]
  \centering
  \resizebox{0.8\textwidth}{!}{ 
  \begin{tabular}{lc|c|c|c}
    \toprule[1.25pt]
    \textbf{Method}
      & \textbf{GT Labels}
      & \textbf{FiNER (Acc$\uparrow$)} 
      & \textbf{Formula (Acc$\uparrow$)}
      & \textbf{Average}\\
    \midrule[1.1pt]
    \rowcolor[rgb]{0.93,0.93,0.93}
    \multicolumn{5}{c}{\texttt{DeepSeek-V3.1 as Base LLM}} \\
    \textcolor{gray}{\texttt{Base LLM}} & & \textcolor{gray}{70.7} & \textcolor{gray}{67.5} & \textcolor{gray}{69.1}\\
    \midrule
    \rowcolor[rgb]{0.93,0.93,0.93}
    \multicolumn{5}{c}{\texttt{Offline Adaptation}} \\
    ICL & \checkmark & $72.3_{\textcolor{scoregreen}{+1.6}}$ & $67.0_{\textcolor{red}{-0.5}}$ & $69.6_{\textcolor{scoregreen}{+0.5}}$ \\
    MIPROv2 & \checkmark & $72.4_{\textcolor{scoregreen}{+1.7}}$ & $69.5_{\textcolor{scoregreen}{+2.0}}$ & $70.9_{\textcolor{scoregreen}{+1.8}}$ \\
    GEPA & \checkmark & $73.5_{\textcolor{scoregreen}{+2.8}}$ & $71.5_{\textcolor{scoregreen}{+4.0}}$ & $72.5_{\textcolor{scoregreen}{+3.4}}$ \\
    ACE & \checkmark & $\textcolor{black}{\mathbf{78.3}}_{\textcolor{scoregreen}{\mathbf{+7.6}}}$ & $\textcolor{black}{\mathbf{85.5}}_{\textcolor{scoregreen}{\mathbf{+18.0}}}$ & $\textcolor{black}{\mathbf{81.9}}_{\textcolor{scoregreen}{\mathbf{+12.8}}}$ \\
    ACE & \xmark & $71.1_{\textcolor{scoregreen}{+0.4}}$ & $83.0_{\textcolor{scoregreen}{+15.5}}$ & $77.1_{\textcolor{scoregreen}{+8.0}}$ \\
    \midrule
    \rowcolor[rgb]{0.93,0.93,0.93}
    \multicolumn{5}{c}{\texttt{Online Adaptation}} \\
    DC (CU) & \checkmark & $74.2_{\textcolor{scoregreen}{+3.5}}$ & $69.5_{\textcolor{scoregreen}{+2.0}}$ & $71.8_{\textcolor{scoregreen}{+2.7}}$ \\
    DC (CU) & \xmark & $68.3_{\textcolor{red}{-2.4}}$ & $62.5_{\textcolor{red}{-5.0}}$ & $65.4_{\textcolor{red}{-3.7}}$ \\
    ACE & \checkmark & $\textcolor{black}{\mathbf{76.7}}_{\textcolor{scoregreen}{\mathbf{+6.0}}}$ & $76.5_{\textcolor{scoregreen}{+9.0}}$ & $\textcolor{black}{\mathbf{76.6}}_{\textcolor{scoregreen}{\mathbf{+7.5}}}$ \\
    ACE & \xmark & $67.3_{\textcolor{red}{-3.4}}$ & $\mathbf{78.5}_{\textcolor{scoregreen}{\mathbf{+11.0}}}$ & $72.9_{\textcolor{scoregreen}{+3.8}}$ \\
    \bottomrule[1.25pt]
  \end{tabular}}
  \caption{\textbf{Results on Financial Analysis Benchmark (DeepSeek-V3.1-671B as the Base LLM).} ``GT labels" indicates whether ground-truth labels are available to the Reflector during adaptation. With GT labels, ACE achieves consistent improvements in both offline and online settings, highlighting the advantage of structured and evolving contexts for domain-specific reasoning. However, we also observe that in the absence of reliable feedback signals (\eg ground-truth labels or execution outcomes), both ACE and other adaptive methods such as Dynamic Cheatsheet may degrade, suggesting that context adaptation depends critically on feedback quality.}
  \label{tab:ds-results}
  \vspace{-5pt}
\end{table*}

\paragraph{Analysis: Finance Benchmark} As shown in Table~\ref{tab:ds-results}, ACE delivers strong improvements on financial analysis benchmarks.  
In the offline setting, when provided with ground-truth answers from the training split, ACE surpasses ICL, MIPROv2, and GEPA by clear margins (an average of 10.9\%), showing that structured and evolving contexts are particularly effective when tasks require precise domain knowledge (\eg financial concepts, XBRL rules) that goes beyond fixed demonstrations or monolithic optimized prompts.  
In the online setting, ACE continues to exceed prior adaptive methods such as DC by an average of 6.2\%, further confirming the benefit of agentic context engineering for accumulating reusable insights across specialized domains. 

Moreover, we also observe that when ground-truth supervision or reliable execution signals are absent, both ACE and DC may degrade in performance.  
In such cases, the constructed context can be polluted by spurious or misleading signals, highlighting a potential limitation of inference-time adaptation without reliable feedback.  
This suggests that while ACE is robust under rich feedback (\eg code execution results or formula correctness in agent tasks), its effectiveness depends on the availability of signals that allow the Reflector and Curator to make sound judgments.  
We return to this limitation in \S\ref{sec:discussion}. 

\paragraph{Analysis: Medical and Text-to-SQL Benchmark} While this subsection focuses on finance as a detailed case study, ACE is not finance-specific: we also see consistent gains on other domain-specific tasks, including medical reasoning and text-to-SQL, suggesting that the same playbook-style context adaptation transfers across domains. 
Full results are reported in Appendix~\S\ref{subsec:streambench-results}.

\subsection{Generalization across LLMs} 
Table~\ref{tab:appworld-results} and Table~\ref{tab:ds-results} use our default backbone (DeepSeek-V3.1), but ACE is not specific to this model. 
We can swap in other LLMs without changing the algorithm or prompts, and still see consistent gains on AppWorld and Finance benchmarks. 
Appendix~\S\ref{subsec:generalization-across-llms} reports full results on GPT-OSS-120B, GPT-5.1, and Llama-3.3-70B-Instruct, where ACE improves over the corresponding base agents or models. 
These results suggest ACE is a generalizable method for test-time context evolution across LLM families.

\subsection{Ablation Study and Sensitivity Analysis}
\label{subsec:results-ablation}

\paragraph{Ablation Study} Table~\ref{tab:ablation} reports ablation studies on AppWorld, analyzing how individual design choices of \name contribute to effective context adaptation. 
We examine three factors: (1) \textit{the Reflector with iterative refinement}, our addition to the agentic framework beyond Dynamic Cheatsheet, (2) \textit{multi-epoch adaptation}, which refines contexts over training samples multiple times, and (3) \textit{offline warmup}, which initializes the context through offline adaptation before online adaptation begins.
Additionally, we study the effect of \emph{incremental context update} and why it is a key enabler for ACE's performance gain in Appendix~\S\ref{subsec:incremental-update-ablation}.

\begin{table*}[!tbhp]
  \centering
  \resizebox{\textwidth}{!}{ 
  \begin{tabular}{lc|cc|cc|c}
    \toprule[1.25pt]
    \multirow{2}{*}{\textbf{Method}} 
      & \multirow{2}{*}{\textbf{GT Labels}}
      & \multicolumn{2}{c|}{\textbf{Test-Normal}} 
      & \multicolumn{2}{c|}{\textbf{Test-Challenge}} 
      & \multirow{2}{*}{\textbf{Average}}\\
    \cmidrule(lr){3-4}\cmidrule(lr){5-6}
      & & \textbf{TGC$\uparrow$} & \textbf{SGC$\uparrow$} & \textbf{TGC$\uparrow$} & \textbf{SGC$\uparrow$} \\
    \midrule[1.1pt]
    \rowcolor[rgb]{0.93,0.93,0.93}
    \multicolumn{7}{c}{\texttt{DeepSeek-V3.1 as Base LLM}} \\
    \textcolor{gray}{\texttt{ReAct}} & & \textcolor{gray}{63.7} & \textcolor{gray}{42.9} & \textcolor{gray}{41.5} & \textcolor{gray}{21.6} & \textcolor{gray}{42.4} \\
    \midrule
    \rowcolor[rgb]{0.93,0.93,0.93}
    \multicolumn{7}{c}{\texttt{Offline Adaptation}} \\
    \texttt{ReAct} + ACE w/o Reflector or multi-epoch & \checkmark  
      & $70.8_{\textcolor{scoregreen}{+7.1}}$ 
      & $55.4_{\textcolor{scoregreen}{+12.5}}$ 
      & $55.9_{\textcolor{scoregreen}{+14.4}}$ 
      & $38.1_{\textcolor{scoregreen}{+17.5}}$
      & $55.1_{\textcolor{scoregreen}{+12.7}}$\\
    \texttt{ReAct} + ACE w/o multi-epoch & \checkmark 
      & $72.0_{\textcolor{scoregreen}{+8.3}}$ 
      & $60.7_{\textcolor{scoregreen}{+17.8}}$ 
      & $54.9_{\textcolor{scoregreen}{+13.4}}$ 
      & $39.6_{\textcolor{scoregreen}{+18.0}}$ 
      & $56.8_{\textcolor{scoregreen}{+14.4}}$\\
    \texttt{ReAct} + ACE & \checkmark 
      & $76.2_{\textcolor{scoregreen}{+12.5}}$ 
      & $64.3_{\textcolor{scoregreen}{+21.4}}$ 
      & $57.3_{\textcolor{scoregreen}{+15.8}}$ 
      & $39.6_{\textcolor{scoregreen}{+18.0}}$ 
      & $59.4_{\textcolor{scoregreen}{+17.0}}$\\
    \midrule
    \rowcolor[rgb]{0.93,0.93,0.93}
    \multicolumn{7}{c}{\texttt{Online Adaptation}} \\
    \texttt{ReAct} + ACE & \xmark 
      & $67.9_{\textcolor{scoregreen}{+4.2}}$ 
      & $51.8_{\textcolor{scoregreen}{+8.9}}$ 
      & $61.4_{\textcolor{scoregreen}{+19.9}}$ 
      & $43.2_{\textcolor{scoregreen}{+21.6}}$ 
      & $56.1_{\textcolor{scoregreen}{+13.7}}$\\
    \texttt{ReAct} + ACE + offline warmup & \xmark 
      & $69.6_{\textcolor{scoregreen}{+5.9}}$ 
      & $53.6_{\textcolor{scoregreen}{+10.7}}$ 
      & $66.0_{\textcolor{scoregreen}{+24.5}}$ 
      & $48.9_{\textcolor{scoregreen}{+27.3}}$ 
      & $59.5_{\textcolor{scoregreen}{+17.1}}$ \\
    \bottomrule[1.25pt]
  \end{tabular}}
  \caption{\textbf{Ablation Studies on AppWorld.} We study how particular design choices of \name (iterative refinement, multi-epoch adaptation, and offline warmup) could help high-quality context adaptation. }
  \label{tab:ablation}
  \vspace{-10pt}
\end{table*}

\paragraph{Robustness to Reflection Quality} \name is robust to reflection quality: it remains effective with a much weaker Reflector and shows only modest additional gains from stronger reflectors, and it degrades gracefully under noisy/harmful reflections, staying above the base model except under fully adversarial updates every iteration. Full experiment results are in Appendix~\S\ref{subsec:robust-reflection}.

\paragraph{Sensitivity to Hyperparameter Choice} \name's gains are stable across a wide range of reasonable hyperparameter settings (\eg Reflector refinement rounds, number of adaptation epochs, and grow-and-refine thresholds): performance changes are modest, and ACE consistently remains above the corresponding baselines. Full discussion and detailed results are reported in Appendix~\S\ref{subsec:hyperparam-sensitivity}.

\subsection{Cost and Speed Analysis}
\label{subsec:results-cost-speed}

Due to its support for incremental, ``delta" context updates and non-LLM-based context merging and de-duplication, \name demonstrates particular advantages in reducing the cost (in terms of the number of rollouts or the amount of dollar cost for token ingestion/generation) and latency of adaptation. 

As examples, on the offline adaptation of AppWorld, ACE achieves 82.3\% reduction in adaptation latency and 75.1\% reduction in the number of rollouts as compared to GEPA (Table \ref{tab:cost-speed}(a)). 
On the online adaptation of FiNER, ACE achieves 91.5\% reduction in adaptation latency and 83.6\% reduction in token dollar cost for token ingestion/generation as compared to DC (Table \ref{tab:cost-speed}(b)).

\begin{table}[!tbhp]
  \centering
  \begin{minipage}[t]{0.45\linewidth}
    \centering
    \resizebox{\linewidth}{!}{%
      \begin{tabular}{@{}lcc@{}}
        \toprule[1.25pt]
        \textbf{Method} & \textbf{Latency (s)$\downarrow$} & \textbf{\# Rollouts$\downarrow$}\\
        \midrule[1.1pt]
        ReAct + GEPA & 53898 & 1434\\
        ReAct + ACE & 9517\textsubscript{\textcolor{red}{(-82.3\%)}} & 357\textsubscript{\textcolor{red}{(-75.1\%)}}\\
        \bottomrule
      \end{tabular}
    }
    \vspace{2pt}\\
    (a) \textbf{Offline} (AppWorld).
  \end{minipage}
  \hfill
  \begin{minipage}[t]{0.43\linewidth}
    \centering
    \resizebox{\linewidth}{!}{%
      \begin{tabular}{@{}lcc@{}}
        \toprule[1.25pt]
        \textbf{Method} & \textbf{Latency (s)$\downarrow$} & \textbf{Token Cost (\$)$\downarrow$}\\
        \midrule[1.1pt]
        DC (CU) & 65104 & 17.7\\
        ACE & 5503\textsubscript{\textcolor{red}{(-91.5\%)}} & 2.9\textsubscript{\textcolor{red}{(-83.6\%)}}\\
        \bottomrule
      \end{tabular}
    }
    \vspace{2pt}\\
    (b) \textbf{Online} (FiNER).
  \end{minipage}

  \vspace{3pt}
  \caption{\textbf{Cost and Speed Analysis.} We measure the context adaptation latency, number of rollouts, and dollar costs of \name against GEPA (offline) and DC (online).}
  \label{tab:cost-speed}
  \vspace{-10pt}
\end{table}

\paragraph{Fine-Grained Cost Analysis} 
We conduct a fine-grained cost analysis of \name and GEPA (as a representative baseline).
On AppWorld, \name is substantially cheaper during offline adaptation, reducing input/output token usage by \textbf{80.8\%/83.6\%} vs.~GEPA: \name avoids GEPA's prompt-validation loop and replaces repeated full rewrites with localized delta updates.
At evaluation time, while \name may use more \emph{raw} input tokens due to a richer playbook, this does not necessarily translate to higher \emph{billed} serving cost because a large fraction of the context is reused by KV caching; we quantify this effect in the next paragraph.
Full results, including a component-wise token breakdown for both methods, are in Appendix~\S\ref{subsec:fine-grained-cost-breakdown}.

\paragraph{KV Cache Reuse: Longer Context $\ne$ Higher Serving Cost} 
Although \name produces longer contexts than methods such as GEPA, this does not translate to linearly higher inference cost or GPU memory usage.  
Modern serving infrastructures are increasingly optimized for long-context workloads through techniques such as the reuse~\citep{gim2024prompt, yao2025cacheblend}, compression~\citep{liu2024kivi, liu2024cachegen}, and offload~\citep{lee2024infinigen, li2025continuum} of KV cache.  
These mechanisms allow frequently reused context segments to be cached locally or remotely, avoiding repetitive and expensive prefill operations.  
Ongoing advances in ML systems suggest that the amortized cost of handling long contexts is likely to decrease, making context-rich approaches like \name increasingly practical in deployment. 
In our prompt-caching study with the OpenAI API (GPT-5.1), we find that \name achieves \emph{high cache reuse}: \textbf{91.8\%} of input tokens are served from cache during evaluation stage, which reduces billed input-token cost by \textbf{82.6\%} relative to counting raw context tokens.
\section{Discussion}
\label{sec:discussion}

\paragraph{Implications for Online and Continual Learning} 
Online and continual learning are key research directions in machine learning for addressing issues like distribution shifts~\citep{koh2021wilds,gulrajani2021domain} and limited training data~\citep{pan2010survey,hutchinson2017overcoming,zhuang2019transfer}.
\name offers a flexible and efficient alternative to conventional model fine-tuning, as adapting contexts is generally cheaper than updating model weights~\citep{brown2020gpt3,lester2021prompttuning,li2021prefixtuning,hu2021lora}.
Moreover, because contexts are human-interpretable, \name enables \textit{selective unlearning}~\citep{cao2015unlearning,bourtoule2021sisa,liu2024rethinkingllmunlearning}, whether due to privacy or legal constraints~\citep{gdpr2016art17,ccpa1798_105}, or when outdated or incorrect information is identified by domain experts.
These are promising directions for future work, where \name could play a central role in advancing continuous and responsible learning.

\paragraph{Limitations and Challenges}  
A limitation of \name is its reliance on a reasonably strong Reflector: if the Reflector fails to extract meaningful insights from generated traces or outcomes, the constructed context may become noisy or even harmful.  
In domain-specific tasks where no model can extract useful insights, the resulting context will naturally lack them. This dependency is similar to Dynamic Cheatsheet~\citep{suzgun2025dynamic}, where the quality of adaptation hinges on the underlying model’s ability to curate memory.  
We also note that not all applications require rich or detailed contexts.  
Tasks like HotPotQA~\citep{yang2018hotpotqa} often benefit more from concise, high-level instructions (\eg how to retrieve and synthesize evidence) than from long contexts.  
Similarly, games with fixed strategies such as Game of 24~\citep{suzgun2025dynamic} may only need a single reusable rule, rendering additional context redundant.  
Overall, \name is most beneficial in settings that demand detailed domain knowledge, complex tool use, or environment-specific strategies that go beyond what is already embedded in model weights or simple system instructions.

\section*{Acknowledgement}

We thank the anonymous reviewers and area chair for their constructive feedback, which improved this paper.
Qizheng Zhang is supported by NSF award CNS-2211384 and DARPA award TFAWI-HR00112520038.
We also thank Lakshya A Agrawal, Xuekai Zhu, Yuhan Liu, Junchen Jiang, and Azalia Mirhoseini for helpful discussions.
\section*{Ethics Statement}

This work does not raise specific ethical concerns. 
Our contributions focus on developing algorithms and system frameworks for effective context adaptation in large language models (LLMs). 
All experiments are conducted on publicly available benchmarks with open-source models, without involving human subjects, sensitive data, or privacy-related information. 
No potential conflicts of interest are present.

\section*{Reproducibility Statement}

Our code is available at \texttt{github.com/ace-agent/ace}.
We provide detailed descriptions of our experimental setup, including datasets, benchmarks, evaluation metrics, baselines, and hyperparameter choices. 
Additional details, such as prompts for large language models and extended experimental settings, are included in the appendix. 
With this information, readers with reasonable computational resources should be able to reproduce our results.

\bibliography{iclr2026_conference}

@article{suzgun2025dynamic,
  title={{Dynamic Cheatsheet: Test-Time Learning with Adaptive Memory}},
  author={Suzgun, Mirac and Yuksekgonul, Mert and Bianchi, Federico and Jurafsky, Dan and Zou, James},
  journal={arXiv preprint arXiv:2504.07952},
  year={2025}
}

@article{agrawal2025gepa,
  title={{GEPA: Reflective Prompt Evolution Can Outperform Reinforcement Learning}},
  author={Agrawal, Lakshya A and Tan, Shangyin and Soylu, Dilara and Ziems, Noah and Khare, Rishi and Opsahl-Ong, Krista and Singhvi, Arnav and Shandilya, Herumb and Ryan, Michael J and Jiang, Meng and others},
  journal={arXiv preprint arXiv:2507.19457},
  year={2025}
}

@inproceedings{wang2024agent,
title={{Agent Workflow Memory}},
author={Zora Zhiruo Wang and Jiayuan Mao and Daniel Fried and Graham Neubig},
booktitle={Forty-second International Conference on Machine Learning (ICML)},
year={2025}
}

@inproceedings{zhang2025cost,
  title={{Agentic Plan Caching: Test-Time Memory for Fast and Cost-Efficient LLM Agents}},
  author={Qizheng Zhang and Michael Wornow and Kunle Olukotun},
  booktitle={The Thirty-ninth Annual Conference on Neural Information Processing Systems (NeurIPS)},
  year={2025}
}

@article{wang2025finlora,
  title={{FinLoRA: Benchmarking LoRA Methods for Fine-Tuning LLMs on Financial Datasets}},
  author={Wang, Dannong and Patel, Jaisal and Zha, Daochen and Yang, Steve Y and Liu, Xiao-Yang},
  journal={arXiv preprint arXiv:2505.19819},
  year={2025}
}

@article{patil2024gorilla,
  title={{Gorilla: Large Language Model Connected with Massive APIs}},
  author={Patil, Shishir G and Zhang, Tianjun and Wang, Xin and Gonzalez, Joseph E},
  journal={Advances in Neural Information Processing Systems (NeurIPS)},
  year={2024}
}

@inproceedings{liu2024cachegen,
  title={{CacheGen: KV Cache Compression and Streaming for Fast Large Language Model Serving}},
  author={Liu, Yuhan and Li, Hanchen and Cheng, Yihua and Ray, Siddhant and Huang, Yuyang and Zhang, Qizheng and Du, Kuntai and Yao, Jiayi and Lu, Shan and Ananthanarayanan, Ganesh and others},
  booktitle={Proceedings of the ACM SIGCOMM 2024 Conference},
  pages={38--56},
  year={2024}
}

@inproceedings{yao2025cacheblend,
  title={{CacheBlend: Fast Large Language Model Serving for RAG with Cached Knowledge Fusion
}},
  author={Yao, Jiayi and Li, Hanchen and Liu, Yuhan and Ray, Siddhant and Cheng, Yihua and Zhang, Qizheng and Du, Kuntai and Lu, Shan and Jiang, Junchen},
  booktitle={Proceedings of the Twentieth European Conference on Computer Systems (EuroSys)},
  pages={94--109},
  year={2025}
}

@article{yang2024swe,
  title={{SWE-agent: Agent-Computer Interfaces Enable Automated Software Engineering}},
  author={Yang, John and Jimenez, Carlos E and Wettig, Alexander and Lieret, Kilian and Yao, Shunyu and Narasimhan, Karthik and Press, Ofir},
  journal={Advances in Neural Information Processing Systems (NeurIPS)},
  year={2024}
}

@article{yuksekgonul2024textgrad,
  title={{Optimizing Generative AI by Backpropagating Language Model Feedback}},
  author={Yuksekgonul, Mert and Bianchi, Federico and Boen, Joseph and Liu, Sheng and Lu, Pan and Huang, Zhi and Guestrin, Carlos and Zou, James},
  journal={Nature},
  volume={639},
  number={8055},
  pages={609--616},
  year={2025},
  publisher={Nature Publishing Group UK London}
}

@article{shinn2023reflexion,
  title={{Reflexion: Language Agents with Verbal Reinforcement Learning}},
  author={Shinn, Noah and Cassano, Federico and Gopinath, Ashwin and Narasimhan, Karthik and Yao, Shunyu},
  journal={Advances in Neural Information Processing Systems (NeurIPS)},
  year={2023}
}

@article{guha2023legalbench,
  title={{LegalBench: A Collaboratively Built Benchmark for Measuring Legal Reasoning in Large Language Models}},
  author={Guha, Neel and Nyarko, Julian and Ho, Daniel and R{\'e}, Christopher and Chilton, Adam and Chohlas-Wood, Alex and Peters, Austin and Waldon, Brandon and Rockmore, Daniel and Zambrano, Diego and others},
  journal={Advances in neural information processing systems (NeurIPS)},
  year={2023}
}

@inproceedings{trivedi2024appworld,
  title={{AppWorld: A Controllable World of Apps and People for Benchmarking Interactive Coding Agents}},
  author={Trivedi, Harsh and Khot, Tushar and Hartmann, Mareike and Manku, Ruskin and Dong, Vinty and Li, Edward and Gupta, Shashank and Sabharwal, Ashish and Balasubramanian, Niranjan},
  booktitle={Proceedings of the 62nd Annual Meeting of the Association for Computational Linguistics (Volume 1: Long Papers)},
  pages={16022--16076},
  year={2024}
}

@article{agarwal2024many,
  title={{Many-Shot In-Context Learning}},
  author={Agarwal, Rishabh and Singh, Avi and Zhang, Lei and Bohnet, Bernd and Rosias, Luis and Chan, Stephanie and Zhang, Biao and Anand, Ankesh and Abbas, Zaheer and Nova, Azade and others},
  journal={Advances in Neural Information Processing Systems (NeurIPS)},
  volume={37},
  pages={76930--76966},
  year={2024}
}

@article{zhou2025agentfly,
  title={{AgentFly: Fine-tuning LLM Agents without Fine-tuning LLMs}},
  author={Zhou, Huichi and Chen, Yihang and Guo, Siyuan and Yan, Xue and Lee, Kin Hei and Wang, Zihan and Lee, Ka Yiu and Zhang, Guchun and Shao, Kun and Yang, Linyi and others},
  journal={arXiv preprint arXiv:2508.16153},
  year={2025}
}

@inproceedings{zhang2024caravan,
  title={{Caravan: Practical Online Learning of In-Network ML Models with Labeling Agents}},
  author={Zhang, Qizheng and Imran, Ali and Bardhi, Enkeleda and Swamy, Tushar and Zhang, Nathan and Shahbaz, Muhammad and Olukotun, Kunle},
  booktitle={18th USENIX Symposium on Operating Systems Design and Implementation (OSDI)},
  pages={325--345},
  year={2024}
}

@article{krause2019dynamic,
  title={{Dynamic Evaluation of Transformer Language Models}},
  author={Krause, Ben and Kahembwe, Emmanuel and Murray, Iain and Renals, Steve},
  journal={arXiv preprint arXiv:1904.08378},
  year={2019}
}

@inproceedings{asai2024self,
  title={{Self-RAG: Learning to Retrieve, Generate, and Critique through Self-Reflection}},
  author={Asai, Akari and Wu, Zeqiu and Wang, Yizhong and Sil, Avirup and Hajishirzi, Hannaneh},
  booktitle={The Twelfth International Conference on Learning Representations (ICLR)},
  year={2024}
}

@inproceedings{ye2023generating,
  title={{Generating Data for Symbolic Language with Large Language Models}},
  author={Ye, Jiacheng and Li, Chengzu and Kong, Lingpeng and Yu, Tao},
  booktitle={Proceedings of the 2023 conference on empirical methods in natural language processing (EMNLP)},
  pages={8418--8443},
  year={2023}
}

@inproceedings{loukas2022finer,
  title={{FiNER: Financial Numeric Entity Recognition for XBRL Tagging}},
  author={Loukas, Lefteris and Fergadiotis, Manos and Chalkidis, Ilias and Spyropoulou, Eirini and Malakasiotis, Prodromos and Androutsopoulos, Ion and Paliouras, Georgios},
  booktitle={Proceedings of the 60th Annual Meeting of the Association for Computational Linguistics (Volume 1: Long Papers)},
  pages={4419--4431},
  year={2022}
}

@inproceedings{yao2023react,
  title={{ReAct: Synergizing Reasoning and Acting in Language Models}},
  author={Yao, Shunyu and Zhao, Jeffrey and Yu, Dian and Du, Nan and Shafran, Izhak and Narasimhan, Karthik and Cao, Yuan},
  booktitle={International Conference on Learning Representations (ICLR)},
  year={2023}
}

@misc{zaharia2024compoundGS,
  author       = {Zaharia, Matei and Khattab, Omar and Chen, Lingjiao and Davis, Jared Quincy and Miller, Heather and Potts, Chris and Zou, James and Carbin, Michael and Frankle, Jonathan and Rao, Naveen and Ghodsi, Ali},
  title        = {{The Shift from Models to Compound AI Systems}},
  howpublished = {\url{https://bair.berkeley.edu/blog/2024/02/18/compound-ai-systems/}},
  year         = {2024},
}

@article{jiang2025putting,
  title={{Putting It All into Context: Simplifying Agents with LCLMs}},
  author={Jiang, Mingjian and Ruan, Yangjun and Lastras, Luis and Kapanipathi, Pavan and Hashimoto, Tatsunori},
  journal={arXiv preprint arXiv:2505.08120},
  year={2025}
}

@misc{DSPyGEPA,
  author       = {DSPy},
  title        = {{dspy.GEPA: Reflective Prompt Optimizer}},
  howpublished = {\url{https://dspy.ai/api/optimizers/GEPA/overview/}},
  note         = {Accessed: 09/2025}
}

@misc{DSPyMIPROv2,
  author       = {DSPy},
  title        = {{dspy.MIPROv2: Multiprompt Instruction PRoposal Optimizer Version 2}},
  howpublished = {\url{https://dspy.ai/api/optimizers/MIPROv2/}},
  note         = {Accessed: 09/2025}
}

@misc{Suzgun2025_DynamicCheatsheet_code,
  author       = {Mirac Suzgun and Mert Yuksekgonul and Federico Bianchi and Dan Jurafsky and James Zou},
  title        = {{Dynamic Cheatsheet: Test-Time Learning with Adaptive Memory}},
  howpublished = {\url{https://github.com/suzgunmirac/dynamic-cheatsheet}},
  note         = {Accessed: 09/2025}
}

@article{deepseekai2024deepseekv3technicalreport,
  title={{DeepSeek-V3 Technical Report}},
  author={Liu, Aixin and Feng, Bei and Xue, Bing and Wang, Bingxuan and Wu, Bochao and Lu, Chengda and Zhao, Chenggang and Deng, Chengqi and Zhang, Chenyu and Ruan, Chong and others},
  journal={arXiv preprint arXiv:2412.19437},
  year={2024}
}

@misc{AppWorldLeaderboard,
  title        = {Leaderboard},
  author       = {AppWorld},
  howpublished = {\url{https://appworld.dev/leaderboard}},
  note         = {Accessed: 09/2025}
}

@inproceedings{opsahl2024optimizing,
  title={{Optimizing Instructions and Demonstrations for Multi-Stage Language Model Programs}},
  author={Opsahl-Ong, Krista and Ryan, Michael J and Purtell, Josh and Broman, David and Potts, Christopher and Zaharia, Matei and Khattab, Omar},
  booktitle={Proceedings of the 2024 Conference on Empirical Methods in Natural Language Processing (EMNLP)},
  pages={9340--9366},
  year={2024}
}

@inproceedings{yang2018hotpotqa,
  title={{HotpotQA: A Dataset for Diverse, Explainable Multi-hop Question Answering}},
  author={Yang, Zhilin and Qi, Peng and Zhang, Saizheng and Bengio, Yoshua and Cohen, William and Salakhutdinov, Ruslan and Manning, Christopher D},
  booktitle={Proceedings of the 2018 conference on empirical methods in natural language processing (EMNLP)},
  pages={2369--2380},
  year={2018}
}

@article{xu2025mem,
  title={{A-Mem: Agentic Memory for LLM Agents}},
  author={Xu, Wujiang and Mei, Kai and Gao, Hang and Tan, Juntao and Liang, Zujie and Zhang, Yongfeng},
  journal={Advances in neural information processing systems (NeurIPS)},
  year={2025}
}

@article{gim2024prompt,
  title={{Prompt Cache: Modular Attention Reuse for Low-Latency Inference}},
  author={Gim, In and Chen, Guojun and Lee, Seung-seob and Sarda, Nikhil and Khandelwal, Anurag and Zhong, Lin},
  journal={Proceedings of Machine Learning and Systems (MLSys)},
  volume={6},
  pages={325--338},
  year={2024}
}

@article{lewis2020retrieval,
  title={{Retrieval-Augmented Generation for Knowledge-Intensive NLP Tasks}},
  author={Lewis, Patrick and Perez, Ethan and Piktus, Aleksandra and Petroni, Fabio and Karpukhin, Vladimir and Goyal, Naman and K{\"u}ttler, Heinrich and Lewis, Mike and Yih, Wen-tau and Rockt{\"a}schel, Tim and others},
  journal={Advances in neural information processing systems (NeurIPS)},
  year={2020}
}

@inproceedings{borgeaud2022improving,
  title={{Improving Language Models by Retrieving from Trillions of Tokens}},
  author={Borgeaud, Sebastian and Mensch, Arthur and Hoffmann, Jordan and Cai, Trevor and Rutherford, Eliza and Millican, Katie and Van Den Driessche, George Bm and Lespiau, Jean-Baptiste and Damoc, Bogdan and Clark, Aidan and others},
  booktitle={International Conference on Machine Learning (ICML)},
  pages={2206--2240},
  year={2022},
  organization={PMLR}
}

@inproceedings{peng2023yarn,
  title={{YaRN: Efficient Context Window Extension of Large Language Models}},
  author={Peng, Bowen and Quesnelle, Jeffrey and Fan, Honglu and Shippole, Enrico},
  booktitle={The Twelfth International Conference on Learning Representations (ICLR)},
  year={2024},
}

@inproceedings{khot2022decomposed,
  title={{Decomposed Prompting: A Modular Approach for Solving Complex Tasks}},
  author={Khot, Tushar and Trivedi, Harsh and Finlayson, Matthew and Fu, Yao and Richardson, Kyle and Clark, Peter and Sabharwal, Ashish},
  booktitle={International Conference on Learning Representations (ICLR)},
  year={2023}
}

@article{wei2022chain,
  title={{Chain-of-Thought Prompting Elicits Reasoning in Large Language Models}},
  author={Wei, Jason and Wang, Xuezhi and Schuurmans, Dale and Bosma, Maarten and Xia, Fei and Chi, Ed and Le, Quoc V and Zhou, Denny and others},
  journal={Advances in neural information processing systems (NeurIPS)},
  volume={35},
  pages={24824--24837},
  year={2022}
}

@inproceedings{wang2022self,
  title={{Self-Consistency Improves Chain of Thought Reasoning in Language Models}},
  author={Wang, Xuezhi and Wei, Jason and Schuurmans, Dale and Le, Quoc and Chi, Ed and Narang, Sharan and Chowdhery, Aakanksha and Zhou, Denny},
  booktitle={The Eleventh International Conference on Learning Representations (ICLR)},
  year={2023}
}

@article{gao2025prompt,
  title={{The Prompt Alchemist: Automated LLM-Tailored Prompt Optimization for Test Case Generation}},
  author={Gao, Shuzheng and Wang, Chaozheng and Gao, Cuiyun and Jiao, Xiaoqian and Chong, Chun Yong and Gao, Shan and Lyu, Michael},
  journal={arXiv preprint arXiv:2501.01329},
  year={2025}
}

@article{chung2025long,
  title={{Is Long Context All You Need? Leveraging LLM's Extended Context for NL2SQL}},
  author={Chung, Yeounoh and Kakkar, Gaurav T and Gan, Yu and Milne, Brenton and {\"O}zcan, Fatma},
  journal={Proceedings of the VLDB Endowment},
  volume={18},
  number={8},
  pages={2735--2747},
  year={2025},
  publisher={VLDB Endowment}
}

@article{chen2025flora,
  title={{Flora: Effortless Context Construction to Arbitrary Length and Scale}},
  author={Chen, Tianxiang and Tan, Zhentao and Bo, Xiaofan and Wu, Yue and Gong, Tao and Chu, Qi and Ye, Jieping and Yu, Nenghai},
  journal={arXiv preprint arXiv:2507.19786},
  year={2025}
}

@article{mao2024lift,
  title={{LIFT: Improving Long Context Understanding Through Long Input Fine-Tuning}},
  author={Mao, Yansheng and Li, Jiaqi and Meng, Fanxu and Xiong, Jing and Zheng, Zilong and Zhang, Muhan},
  journal={arXiv preprint arXiv:2412.13626},
  year={2024}
}

@inproceedings{zhang2025adaptive,
  title={{Adaptive Self-Improvement LLM Agentic System for ML Library Development}},
  author={Genghan Zhang and Weixin Liang and Olivia Hsu and Kunle Olukotun},
  booktitle={Forty-second International Conference on Machine Learning (ICML)},
  year={2025}
}

@inproceedings{lee2024infinigen,
  title={{InfiniGen: Efficient Generative Inference of Large Language Models with Dynamic KV Cache Management}},
  author={Lee, Wonbeom and Lee, Jungi and Seo, Junghwan and Sim, Jaewoong},
  booktitle={18th USENIX Symposium on Operating Systems Design and Implementation (OSDI 24)},
  pages={155--172},
  year={2024}
}

@inproceedings{liu2024kivi,
  title={{KIVI: A Tuning-Free Asymmetric 2bit Quantization for KV Cache}},
  author={Liu, Zirui and Yuan, Jiayi and Jin, Hongye and Zhong, Shaochen and Xu, Zhaozhuo and Braverman, Vladimir and Chen, Beidi and Hu, Xia},
  booktitle={International Conference on Machine Learning (ICML)},
  year={2024},
}

@article{marreed2025towards,
  title={{Towards Enterprise-Ready Computer Using Generalist Agent}},
  author={Marreed, Sami and Oved, Alon and Yaeli, Avi and Shlomov, Segev and Levy, Ido and Akrabi, Offer and Sela, Aviad and Adi, Asaf and Mashkif, Nir},
  journal={arXiv preprint arXiv:2503.01861},
  year={2025}
}

@inproceedings{liu2025selfelicit,
  title={{SelfElicit: Your Language Model Secretly Knows Where is the Relevant Evidence}},
  author={Liu, Zhining and Amjad, Rana Ali and Adkathimar, Ravinarayana and Wei, Tianxin and Tong, Hanghang},
  booktitle={Proceedings of the 63rd Annual Meeting of the Association for Computational Linguistics (Volume 1: Long Papers)},
  pages={9153--9173},
  year={2025}
}

@inproceedings{koh2021wilds,
  title={{WILDS: A Benchmark of in-the-Wild Distribution Shifts}},
  author={Koh, Pang Wei and Sagawa, Shiori and Marklund, Henrik and Xie, Sang Michael and Zhang, Marvin and Balsubramani, Akshay and Hu, Weihua and Yasunaga, Michihiro and Phillips, Richard Lanas and Gao, Irena and others},
  booktitle={International conference on machine learning (ICML)},
  year={2021}
}

@inproceedings{gulrajani2021domain,
  title={{In Search of Lost Domain Generalization}},
  author={Gulrajani, Ishaan and Lopez-Paz, David},
  booktitle={International Conference on Learning Representations (ICLR)},
  year={2021}
}

@article{pan2010survey,
  title={{A Survey on Transfer Learning}},
  author={Pan, Sinno Jialin and Yang, Qiang},
  journal={IEEE Transactions on Knowledge and Data Engineering},
  volume={22},
  number={10},
  pages={1345--1359},
  year={2010}
}

@article{zhuang2019transfer,
  title={{A Comprehensive Survey on Transfer Learning}},
  author={Zhuang, Fuzhen and Qi, Zhiyuan and Duan, Keyu and Xi, Dongbo and Zhu, Yongchun and Zhu, Hengshu and Xiong, Hui and He, Qing},
  journal={arXiv:1911.02685},
  year={2019}
}

@article{brown2020gpt3,
  title={{Language Models are Few-Shot Learners}},
  author={Brown, Tom and Mann, Benjamin and Ryder, Nick and Subbiah, Melanie and Kaplan, Jared D and Dhariwal, Prafulla and Neelakantan, Arvind and Shyam, Pranav and Sastry, Girish and Askell, Amanda and others},
  journal={Advances in Neural Information Processing Systems (NeurIPS)},
  volume={33},
  pages={1877--1901},
  year={2020}
}

@inproceedings{lester2021prompttuning,
  title={{The Power of Scale for Parameter-Efficient Prompt Tuning}},
  author={Lester, Brian and Al-Rfou, Rami and Constant, Noah},
  booktitle={Proceedings of the 2021 conference on empirical methods in natural language processing (EMNLP)},
  pages={3045--3059},
  year={2021}
}

@inproceedings{li2021prefixtuning,
  title={{Prefix-Tuning: Optimizing Continuous Prompts for Generation}},
  author={Li, Xiang Lisa and Liang, Percy},
  booktitle={Proceedings of the 59th Annual Meeting of the Association for Computational Linguistics and the 11th International Joint Conference on Natural Language Processing (Volume 1: Long Papers)},
  pages={4582--4597},
  year={2021}
}

@inproceedings{hu2021lora,
  title={{LoRA: Low-Rank Adaptation of Large Language Models}},
  author={Hu, Edward J. and Shen, Yelong and Wallis, Phillip and Allen-Zhu, Zeyuan and Li, Yuanzhi and Wang, Shean and Wang, Lu and Chen, Weizhu},
  booktitle={The Tenth International Conference on Learning Representations (ICLR)},
  year={2022}
}

@inproceedings{cao2015unlearning,
  title={{Towards Making Systems Forget with Machine Unlearning}},
  author={Cao, Yinzhi and Yang, Junfeng},
  booktitle={IEEE Symposium on Security and Privacy  (IEEE S\&P)},
  year={2015}
}

@inproceedings{bourtoule2021sisa,
  title={{Machine Unlearning}},
  author={Bourtoule, Lucas and Chandrasekaran, Varun and Choquette-Choo, Christopher and Jia, Hengrui and Travers, Adelin and Zhang, Baiwu and Lie, David and Papernot, Nicolas},
  booktitle={IEEE Symposium on Security and Privacy (IEEE S\&P)},
  year={2021}
}

@article{liu2024rethinkingllmunlearning,
  title={{Rethinking Machine Unlearning for Large Language Models}},
  author={Liu, Sijia and Yao, Yuanshun and Jia, Jinghan and Casper, Stephen and Baracaldo, Nathalie and Hase, Peter and Yao, Yuguang and Liu, Chris Yuhao and Xu, Xiaojun and Li, Hang and others},
  journal={Nature Machine Intelligence},
  volume={7},
  number={2},
  pages={181--194},
  year={2025},
  publisher={Nature Publishing Group UK London}
}

@misc{gdpr2016art17,
  title={{General Data Protection Regulation (GDPR) Article 17: Right to Erasure}},
  howpublished={EU Regulation 2016/679},
  year={2016}
}

@misc{ccpa1798_105,
  title={{California Consumer Privacy Act, Civil Code \S1798.105: Right to Delete}},
  howpublished={State of California Civil Code},
  year={2018}
}

@article{hutchinson2017overcoming,
  title={{Overcoming Data Scarcity with Transfer Learning}},
  author={Hutchinson, Maxwell L and Antono, Erin and Gibbons, Brenna M and Paradiso, Sean and Ling, Julia and Meredig, Bryce},
  journal={arXiv preprint arXiv:1711.05099},
  year={2017}
}

@article{mang2025frontiercs,
  title={{FrontierCS: Evolving Challenges for Evolving Intelligence}},
  author={Mang, Qiuyang and Chai, Wenhao and Li, Zhifei and Mao, Huanzhi and Zhou, Shang and Du, Alexander and Li, Hanchen and Liu, Shu and Chen, Edwin and Wang, Yichuan and others},
  journal={arXiv preprint arXiv:2512.15699},
  year={2025}
}

@article{zhang2025accelopt,
  title={{AccelOpt: A Self-Improving LLM Agentic System for AI Accelerator Kernel Optimization}},
  author={Zhang, Genghan and Zhu, Shaowei and Wei, Anjiang and Song, Zhenyu and Nie, Allen and Jia, Zhen and Vijaykumar, Nandita and Wang, Yida and Olukotun, Kunle},
  journal={arXiv preprint arXiv:2511.15915},
  year={2025}
}

@article{li2025continuum,
  title={{Continuum: Efficient and Robust Multi-Turn LLM Agent Scheduling with KV Cache Time-to-Live}},
  author={Li, Hanchen and Mang, Qiuyang and He, Runyuan and Zhang, Qizheng and Mao, Huanzhi and Chen, Xiaokun and Zhou, Hangrui and Cheung, Alvin and Gonzalez, Joseph and Stoica, Ion},
  journal={arXiv preprint arXiv:2511.02230},
  year={2025}
}

@article{wu2024streambench,
  title={{StreamBench: Towards Benchmarking Continuous Improvement of Language Agents}},
  author={Wu, Cheng-Kuang and Tam, Zhi R and Lin, Chieh-Yen and Chen, Yun-Nung and Lee, Hung-yi},
  journal={Advances in Neural Information Processing Systems (NeurIPS)},
  year={2024}
}

@article{fansi2022ddxplus,
  title={{DDXPlus: A New Dataset For Automatic Medical Diagnosis}},
  author={Fansi Tchango, Arsene and Goel, Rishab and Wen, Zhi and Martel, Julien and Ghosn, Joumana},
  journal={Advances in neural information processing systems (NeurIPS)},
  year={2022}
}

@article{li2023can,
  title={{Can LLM Already Serve as A Database Interface? A BIg Bench for Large-Scale Database Grounded Text-to-SQLs}},
  author={Li, Jinyang and Hui, Binyuan and Qu, Ge and Yang, Jiaxi and Li, Binhua and Li, Bowen and Wang, Bailin and Qin, Bowen and Geng, Ruiying and Huo, Nan and others},
  journal={Advances in Neural Information Processing Systems (NeurIPS)},
  year={2023}
}

@article{zheng2023judging,
  title={{Judging LLM-as-a-Judge with MT-Bench and Chatbot Arena}},
  author={Zheng, Lianmin and Chiang, Wei-Lin and Sheng, Ying and Zhuang, Siyuan and Wu, Zhanghao and Zhuang, Yonghao and Lin, Zi and Li, Zhuohan and Li, Dacheng and Xing, Eric and others},
  journal={Advances in neural information processing systems (NeurIPS)},
  volume={36},
  pages={46595--46623},
  year={2023}
}

@misc{gpt4omini,
  author       = {OpenAI},
  title        = {{GPT‑4o mini: Advancing Cost-Efficient Intelligence}},
  howpublished = {\url{https://openai.com/index/gpt-4o-mini-advancing-cost-efficient-intelligence/}},
  year         = {2024},
}

@inproceedings{zhou2024defending,
  title={{Defending Jailbreak Prompts via In-Context Adversarial Game}},
  author={Zhou, Yujun and Han, Yufei and Zhuang, Haomin and Guo, Kehan and Liang, Zhenwen and Bao, Hongyan and Zhang, Xiangliang},
  booktitle={Proceedings of the 2024 Conference on Empirical Methods in Natural Language Processing (EMNLP)},
  pages={20084--20105},
  year={2024}
}
\bibliographystyle{iclr2026_conference}

\appendix
\newpage
\section{Extended Results}

\subsection{Generalization across Different LLMs}
\label{subsec:generalization-across-llms}

\name is a model-agnostic framework that operates on execution traces and contextual deltas, and does not rely on any architectural or training-specific features of DeepSeek-V3.1 (the default backbone we use in the main text). 
We evaluated \name with three additional models of varying size, cost, and capability: GPT-OSS-120B (Table~\ref{tab:appworld-results-gpt-oss-120b} and \ref{tab:ds-results-gpt-oss-120b}), GPT-5.1 (Table~\ref{tab:appworld-results-gpt51} and \ref{tab:ds-results-gpt51}), and Llama-3.3-70B-Instruct (Table~\ref{tab:ds-results-llama70b}). 
In each case, the Generator, Reflector, and Curator were all switched to the new model without changes to the algorithm.

\paragraph{Analysis} Across all four LLM families we tested (DeepSeek-V3.1, GPT-OSS-120B, GPT-5.1, and Llama-3.3-70B-Instruct), \name consistently improves performance over both the base LLM/agent, GEPA, and other baselines, often by 5 to 12 points depending on the task and supervision setting. The relative gains of \name remain stable even when switching to models that differ significantly in size, cost, and training recipe, and \name delivers benefits with or without ground-truth labels, validating the robustness of its context adaptation mechanism. The online variant reliably achieves the strongest performance across all models. 

We note that the magnitude of improvement can vary across model families: for example, Llama-3.3-70B-Instruct shows smaller gains compared to GPT-5.1 or GPT-OSS-120B. This is expected since \name relies on the quality of intermediate reflections and calibrations, and smaller or weaker models naturally generate noisier feedback. Even in these cases, however, \name remains beneficial, demonstrating that the framework is broadly applicable while still reflecting the inherent capability limits of the underlying LLM, as we discussed in the “Limitations and Challenges” paragraph in \S\ref{sec:discussion}.

\begin{table*}[!tbhp]
  \centering
  \resizebox{\textwidth}{!}{ 
  \begin{tabular}{lc|cc|cc|c}
    \toprule[1.25pt]
    \multirow{2}{*}{\textbf{Method}} 
      & \multirow{2}{*}{\textbf{GT Labels}}
      & \multicolumn{2}{c|}{\textbf{Test-Normal}} 
      & \multicolumn{2}{c|}{\textbf{Test-Challenge}} 
      & \multirow{2}{*}{\textbf{Average}}\\
    \cmidrule(lr){3-4}\cmidrule(lr){5-6}
      & & \textbf{TGC$\uparrow$} & \textbf{SGC$\uparrow$} & \textbf{TGC$\uparrow$} & \textbf{SGC$\uparrow$} \\
    \midrule[1.1pt]
    \rowcolor[rgb]{0.93,0.93,0.93}
    \multicolumn{7}{c}{\texttt{GPT-OSS-120B as Base LLM}} \\
    \textcolor{gray}{\texttt{ReAct}} & & \textcolor{gray}{54.8} & \textcolor{gray}{33.9} & \textcolor{gray}{34.5} & \textcolor{gray}{15.1} & \textcolor{gray}{34.6} \\
    \midrule
    \rowcolor[rgb]{0.93,0.93,0.93}
    \multicolumn{7}{c}{\texttt{Offline Adaptation}} \\
    \texttt{ReAct} + GEPA & \checkmark & $56.0_{\textcolor{scoregreen}{+1.2}}$ & $33.9_{\textcolor{gray}{+0.0}}$ & $40.1_{\textcolor{scoregreen}{+5.6}}$ & $20.9_{\textcolor{scoregreen}{+5.8}}$ & $37.7_{\textcolor{scoregreen}{+3.1}}$\\
    \texttt{ReAct} + \name & \checkmark 
      & $\textcolor{black}{\mathbf{61.3}}_{\textcolor{scoregreen}{\mathbf{+6.5}}}$ 
      & $\textcolor{black}{39.3}_{\textcolor{scoregreen}{\mathbf{+5.4}}}$ 
      & $\textcolor{black}{\mathbf{40.3}}_{\textcolor{scoregreen}{\mathbf{+5.8}}}$ 
      & $\textcolor{black}{\mathbf{20.9}}_{\textcolor{scoregreen}{\mathbf{+5.8}}}$ 
      & $\textcolor{black}{\mathbf{40.5}}_{\textcolor{scoregreen}{\mathbf{+5.9}}}$\\
    \texttt{ReAct} + \name & \xmark 
      & $58.3_{\textcolor{scoregreen}{+3.5}}$ 
      & $\textcolor{black}{\mathbf{41.1}}_{\textcolor{scoregreen}{\mathbf{+7.2}}}$ 
      & $39.6_{\textcolor{scoregreen}{+5.1}}$ 
      & $18.7_{\textcolor{scoregreen}{+3.6}}$ 
      & $39.4_{\textcolor{scoregreen}{+4.8}}$\\
    \midrule
    \rowcolor[rgb]{0.93,0.93,0.93}
    \multicolumn{7}{c}{\texttt{Online Adaptation}} \\
    \texttt{ReAct} + DC (CU) & \xmark 
      & $49.4_{\textcolor{red}{-5.4}}$ 
      & $33.9_{\textcolor{gray}{+0.0}}$ 
      & $30.8_{\textcolor{red}{-3.7}}$ 
      & $18.2_{\textcolor{scoregreen}{+3.1}}$ 
      & $33.1_{\textcolor{red}{-1.5}}$\\
    \texttt{ReAct} + \name & \xmark 
      & $\textcolor{black}{\mathbf{60.7}}_{\textcolor{scoregreen}{\mathbf{+5.9}}}$ 
      & $44.6_{\textcolor{scoregreen}{+10.7}}$ 
      & $\textcolor{black}{\mathbf{43.2}}_{\textcolor{scoregreen}{\mathbf{+8.7}}}$ 
      & $\textcolor{black}{\mathbf{20.1}}_{\textcolor{scoregreen}{\mathbf{+5.0}}}$ 
      & $\textcolor{black}{\mathbf{42.2}}_{\textcolor{scoregreen}{\mathbf{+7.6}}}$\\
    \bottomrule[1.25pt]
  \end{tabular}}
  \caption{\textbf{Results on the AppWorld Agent Benchmark (GPT-OSS-120B as the Base LLM).} ``GT labels" indicates whether ground-truth labels are available to the Reflector during adaptation. We evaluate the \name framework against multiple baselines on top of the official \texttt{ReAct} implementation, both for offline and online context adaptation.}
  \label{tab:appworld-results-gpt-oss-120b}
\end{table*}

\begin{table}[t!]
  \centering
  \resizebox{0.75\linewidth}{!}{ 
  \begin{tabular}{lc|cc|c}
    \toprule[1.25pt]
    \multirow{2}{*}{\textbf{Method}} 
      & \multirow{2}{*}{\textbf{GT Labels}}
      & \multicolumn{2}{c|}{\textbf{Test-Normal}} 
      & \multirow{2}{*}{\textbf{Average}}\\
    \cmidrule(lr){3-4}
      & & \textbf{TGC$\uparrow$} & \textbf{SGC$\uparrow$} \\
    \midrule[1.1pt]
    \rowcolor[rgb]{0.93,0.93,0.93}
    \multicolumn{5}{c}{\texttt{GPT-5.1 as Base LLM}} \\
    \textcolor{gray}{\texttt{ReAct}} & & \textcolor{gray}{61.9} & \textcolor{gray}{46.4} & \textcolor{gray}{54.2} \\
    \midrule
    \rowcolor[rgb]{0.93,0.93,0.93}
    \multicolumn{5}{c}{\texttt{Offline Adaptation}} \\
    \texttt{ReAct} + GEPA & \checkmark 
      & $64.3_{\textcolor{scoregreen}{+2.4}}$ 
      & $48.2_{\textcolor{scoregreen}{+1.8}}$ 
      & $56.2_{\textcolor{scoregreen}{+2.0}}$ \\
    \texttt{ReAct} + \name & \checkmark 
      & $66.7_{\textcolor{scoregreen}{+4.8}}$ 
      & $53.6_{\textcolor{scoregreen}{+7.2}}$ 
      & $60.2_{\textcolor{scoregreen}{+6.0}}$ \\
    \texttt{ReAct} + \name & \xmark 
      & $\textcolor{black}{\mathbf{67.3}}_{\textcolor{scoregreen}{\mathbf{+5.4}}}$ 
      & $\textcolor{black}{\mathbf{55.4}}_{\textcolor{scoregreen}{\mathbf{+9.0}}}$ 
      & $\textcolor{black}{\mathbf{61.3}}_{\textcolor{scoregreen}{\mathbf{+7.1}}}$ \\
    \midrule
    \rowcolor[rgb]{0.93,0.93,0.93}
    \multicolumn{5}{c}{\texttt{Online Adaptation}} \\
    \texttt{ReAct} + DC (CU) & \xmark 
      & $62.5_{\textcolor{scoregreen}{+0.6}}$ 
      & $55.4_{\textcolor{scoregreen}{+9.0}}$ 
      & $58.9_{\textcolor{scoregreen}{+4.7}}$ \\
    \texttt{ReAct} + \name & \xmark 
      & $\textcolor{black}{\mathbf{72.6}}_{\textcolor{scoregreen}{\mathbf{+10.7}}}$ 
      & $\textcolor{black}{\mathbf{58.9}}_{\textcolor{scoregreen}{\mathbf{+12.5}}}$ 
      & $\textcolor{black}{\mathbf{65.8}}_{\textcolor{scoregreen}{\mathbf{+11.6}}}$ \\
    \bottomrule[1.25pt]
  \end{tabular}}
  \caption{\textbf{Results on the AppWorld Agent Benchmark (GPT-5.1 as the Base LLM).} ``GT labels" indicates whether ground-truth labels are available to the Reflector during adaptation. We evaluate the \name framework against multiple baselines on top of the official \texttt{ReAct} implementation, both for offline and online context adaptation.}
  \label{tab:appworld-results-gpt51}
\end{table}

\begin{table*}[t!]
  \centering
  \resizebox{0.8\textwidth}{!}{ 
  \begin{tabular}{lc|c|c|c}
    \toprule[1.25pt]
    \textbf{Method}
      & \textbf{GT Labels}
      & \textbf{FiNER (Acc$\uparrow$)} 
      & \textbf{Formula (Acc$\uparrow$)}
      & \textbf{Average}\\
    \midrule[1.1pt]
    \rowcolor[rgb]{0.93,0.93,0.93}
    \multicolumn{5}{c}{\texttt{GPT-OSS-120B as Base LLM}} \\
    \textcolor{gray}{\texttt{Base LLM}} & & \textcolor{gray}{66.6} & \textcolor{gray}{71.5} & \textcolor{gray}{69.1}\\
    \midrule
    \rowcolor[rgb]{0.93,0.93,0.93}
    \multicolumn{5}{c}{\texttt{Offline Adaptation}} \\
    GEPA & \checkmark & $67.9_{\textcolor{scoregreen}{+1.3}}$ & $71.5_{\textcolor{scoregreen}{+0.0}}$ & $69.7_{\textcolor{scoregreen}{+0.6}}$ \\
    \name & \checkmark & $\textcolor{black}{\mathbf{73.8}}_{\textcolor{scoregreen}{\mathbf{+7.2}}}$ & $\textcolor{black}{\mathbf{88.5}}_{\textcolor{scoregreen}{\mathbf{+17.0}}}$ & $\textcolor{black}{\mathbf{81.2}}_{\textcolor{scoregreen}{\mathbf{+12.1}}}$ \\
    \name & \xmark & $69.7_{\textcolor{scoregreen}{+3.1}}$ & $84.0_{\textcolor{scoregreen}{+12.5}}$ & $76.9_{\textcolor{scoregreen}{+7.8}}$ \\
    \midrule
    \rowcolor[rgb]{0.93,0.93,0.93}
    \multicolumn{5}{c}{\texttt{Online Adaptation}} \\
    DC & \xmark & $55.8_{\textcolor{red}{-10.8}}$ & $66.0_{\textcolor{red}{-5.5}}$ & $60.9_{\textcolor{red}{-8.2}}$ \\
    \name & \xmark & $\mathbf{70.5}_{\textcolor{scoregreen}{\mathbf{+3.9}}}$ & $\mathbf{85.0}_{\textcolor{scoregreen}{\mathbf{+13.5}}}$ & $\mathbf{77.8}_{\textcolor{scoregreen}{\mathbf{+8.7}}}$ \\
    \bottomrule[1.25pt]
  \end{tabular}}
  \caption{\textbf{Results on Financial Analysis Benchmark (GPT-OSS-120B as the Base LLM).} ``GT labels" indicates whether ground-truth labels are available to the Reflector during adaptation.}
  \label{tab:ds-results-gpt-oss-120b}
\end{table*}

\begin{table*}[t!]
  \centering
  \resizebox{0.8\textwidth}{!}{ 
  \begin{tabular}{lc|c|c|c}
    \toprule[1.25pt]
    \textbf{Method}
      & \textbf{GT Labels}
      & \textbf{FiNER (Acc$\uparrow$)} 
      & \textbf{Formula (Acc$\uparrow$)}
      & \textbf{Average}\\
    \midrule[1.1pt]
    \rowcolor[rgb]{0.93,0.93,0.93}
    \multicolumn{5}{c}{\texttt{GPT-5.1 as Base LLM}} \\
    \textcolor{gray}{\texttt{Base LLM}} & & \textcolor{gray}{73.5} & \textcolor{gray}{73.0} & \textcolor{gray}{73.3}\\
    \midrule
    \rowcolor[rgb]{0.93,0.93,0.93}
    \multicolumn{5}{c}{\texttt{Offline Adaptation}} \\
    GEPA & \checkmark & $74.5_{\textcolor{scoregreen}{+1.0}}$ & $73.0_{\textcolor{scoregreen}{+0.0}}$ & $73.8_{\textcolor{scoregreen}{+0.5}}$ \\
    \name & \checkmark & $\textcolor{black}{\mathbf{81.0}}_{\textcolor{scoregreen}{\mathbf{+7.5}}}$ & $\textcolor{black}{\mathbf{84.5}}_{\textcolor{scoregreen}{\mathbf{+11.5}}}$ & $\textcolor{black}{\mathbf{82.8}}_{\textcolor{scoregreen}{\mathbf{+9.5}}}$ \\
    \name & \xmark & $78.2_{\textcolor{scoregreen}{+4.7}}$ & $76.5_{\textcolor{scoregreen}{+3.5}}$ & $77.4_{\textcolor{scoregreen}{+4.1}}$ \\
    \midrule
    \rowcolor[rgb]{0.93,0.93,0.93}
    \multicolumn{5}{c}{\texttt{Online Adaptation}} \\
    DC & \xmark & $70.0_{\textcolor{red}{-3.5}}$ & $69.0_{\textcolor{red}{-4.0}}$ & $69.5_{\textcolor{red}{-3.8}}$ \\
    \name & \xmark & $\mathbf{75.2}_{\textcolor{scoregreen}{\mathbf{+1.7}}}$ & $\mathbf{79.0}_{\textcolor{scoregreen}{\mathbf{+6.0}}}$ & $\mathbf{77.1}_{\textcolor{scoregreen}{\mathbf{+3.8}}}$ \\
    \bottomrule[1.25pt]
  \end{tabular}}
  \caption{\textbf{Results on Financial Analysis Benchmark (GPT-5.1 as the Base LLM).} ``GT labels" indicates whether ground-truth labels are available to the Reflector during adaptation.}
  \label{tab:ds-results-gpt51}
\end{table*}

\begin{table}[t!]
  \centering
  \resizebox{0.55\textwidth}{!}{ 
  \begin{tabular}{lc|c}
    \toprule[1.25pt]
    \textbf{Method}
      & \textbf{GT Labels}
      & \textbf{FiNER (Acc$\uparrow$)} \\
    \midrule[1.1pt]
    \rowcolor[rgb]{0.93,0.93,0.93}
    \multicolumn{3}{c}{\texttt{Llama-3.3-70B-Instruct as Base LLM}} \\
    \textcolor{gray}{\texttt{Base LLM}} & & \textcolor{gray}{62.5} \\
    \midrule

    \rowcolor[rgb]{0.93,0.93,0.93}
    \multicolumn{3}{c}{\texttt{Offline Adaptation}} \\
    GEPA & \checkmark & $59.41_{\textcolor{red}{-3.09}}$ \\
    \name & \checkmark & $\textcolor{black}{\mathbf{64.9}}_{\textcolor{scoregreen}{\mathbf{+2.4}}}$ \\
    \name & \xmark & $64.2_{\textcolor{scoregreen}{+1.7}}$ \\
    \midrule

    \rowcolor[rgb]{0.93,0.93,0.93}
    \multicolumn{3}{c}{\texttt{Online Adaptation}} \\
    DC & \xmark & $59.0_{\textcolor{red}{-3.5}}$ \\
    \name & \xmark & $\mathbf{63.6}_{\textcolor{scoregreen}{\mathbf{+1.1}}}$ \\
    \bottomrule[1.25pt]
  \end{tabular}}
  \caption{\textbf{Results on Financial Analysis Benchmark (Llama-3.3-70B-Instruct as the Base LLM).} ``GT labels" indicates whether ground-truth labels are available to the Reflector during adaptation.}
  \label{tab:ds-results-llama70b}
\end{table}

\subsection{Beyond Finance: Additional Domain Tasks}
\label{subsec:streambench-results}

We evaluate \name on two additional domain tasks from StreamBench~\citep{wu2024streambench} under the non-streaming setting (\ie offline adaptation): DDXPlus~\citep{fansi2022ddxplus} for medical reasoning (Table~\ref{tab:ddxplus-streambench}) and BIRD-SQL~\citep{li2023can} for Text-to-SQL generation (Table~\ref{tab:birdsql-streambench}).
For \name, we perform offline adaptation using 1000 randomly sampled training examples.
For GEPA, we use the same 1000 examples for training, and reserve a separate validation set of 500 examples for BIRD-SQL or 372 examples for DDXPlus (StreamBench provides 1372 train/val examples for DDXPlus in total).
All other settings follow the main-text configuration unless stated otherwise.

\paragraph{Analysis} On DDXPlus, \name substantially improves over the base LLM, rising from 75.2 to 90.2 accuracy (\textbf{+15.0}). 
In contrast, GEPA yields a much smaller gain (76.4, \textbf{+1.2}). 
This suggests that \name's test-time evolving context transfers well to multi-step, domain-heavy diagnostic reasoning.
On BIRD-SQL, \name also improves consistently over the base model on all splits, achieving better overall average (52.9, \textbf{+5.1}). 
The gains are driven mainly by the Simple subset (53.5, \textbf{+7.1}).
While GEPA yields larger gains on Moderate and Challenging, \name still improves over the base model on both splits. 
Overall, these results indicate that \name generalizes beyond finance to both knowledge-intensive reasoning and structured code generation tasks.

\begin{table}[t!]
  \centering
  \resizebox{0.42\textwidth}{!}{
  \begin{tabular}{l|c}
    \toprule[1.25pt]
    \textbf{Method} & \textbf{Accuracy ($\uparrow$)} \\
    \midrule[1.1pt]
    \rowcolor[rgb]{0.93,0.93,0.93}
    \multicolumn{2}{c}{\texttt{DeepSeek-V3.1 as Base LLM}} \\
    \textcolor{gray}{\texttt{Base LLM}} & \textcolor{gray}{75.2} \\
    \midrule
    \rowcolor[rgb]{0.93,0.93,0.93}
    \multicolumn{2}{c}{\texttt{Offline Adaptation}} \\
    GEPA & $76.4_{\textcolor{scoregreen}{+1.2}}$ \\
    \name  & $\textcolor{black}{\mathbf{90.2}}_{\textcolor{scoregreen}{\mathbf{+15.0}}}$ \\
    \bottomrule[1.25pt]
  \end{tabular}}
  \caption{\textbf{Results on Medical Reasoning Benchmark (DeepSeek-V3.1-671B as the Base LLM).} We use DDXPlus from StreamBench.}
  \label{tab:ddxplus-streambench}
\end{table}

\begin{table}[t!]
  \centering
  \resizebox{0.75\textwidth}{!}{
  \begin{tabular}{l|c c c|c}
    \toprule[1.25pt]
    \textbf{Method} & \textbf{Simple ($\uparrow$)} & \textbf{Moderate ($\uparrow$)} & \textbf{Challenging ($\uparrow$)} & \textbf{Average} \\
    \midrule[1.1pt]
    \rowcolor[rgb]{0.93,0.93,0.93}
    \multicolumn{5}{c}{\texttt{DeepSeek-V3.1 as Base LLM}} \\
    \textcolor{gray}{\texttt{Base LLM}} & \textcolor{gray}{46.4} & \textcolor{gray}{48.2} & \textcolor{gray}{55.1} & \textcolor{gray}{47.8} \\
    \midrule
    \rowcolor[rgb]{0.93,0.93,0.93}
    \multicolumn{5}{c}{\texttt{Offline Adaptation}} \\
    GEPA & $51.6_{\textcolor{scoregreen}{+5.2}}$ & $\mathbf{51.9}_{\textcolor{scoregreen}{\mathbf{+3.7}}}$ & $\mathbf{57.2}_{\textcolor{scoregreen}{\mathbf{+2.1}}}$ & $52.2_{\textcolor{scoregreen}{+4.4}}$ \\
    \name  & $\mathbf{53.5}_{\textcolor{scoregreen}{\mathbf{+7.1}}}$ & $50.7_{\textcolor{scoregreen}{+2.5}}$ & $56.6_{\textcolor{scoregreen}{+1.5}}$ & $\mathbf{52.9}_{\textcolor{scoregreen}{\mathbf{+5.1}}}$ \\
    \bottomrule[1.25pt]
  \end{tabular}}
  \caption{\textbf{Results on Text-to-SQL Benchmark (DeepSeek-V3.1-671B as the Base LLM).} We use BIRD-SQL from StreamBench.}
  \label{tab:birdsql-streambench}
\end{table}

\subsection{Fine-Grained Cost Analysis}
\label{subsec:fine-grained-cost-breakdown}

We perform a fine-grained cost analysis of \name and GEPA for both the \emph{adaptation} and \emph{evaluation} stages (offline adaptation setting), using AppWorld as a representative application.
For \name, we run offline adaptation with 1 epoch and 1 reflector refinement round.
While increasing the number of epochs or refinement rounds will increase cost, the qualitative trends below should remain: \name avoids expensive validation-time re-evaluation and performs localized updates rather than repeated full rewrites.
For GEPA, we use the official DSPy implementation~\citep{DSPyMIPROv2} with \verb|auto="heavy"| to maximize optimization strength.

\paragraph{Adaptation Stage}
Across adaptation (Table~\ref{tab:cost-adapt-aggregate} and Table~\ref{tab:cost-adapt-average}), \name reduces input-token usage by 80.8\% relative to GEPA (204.1M $\rightarrow$ 39.3M) and output-token usage by 83.6\% (1.87M $\rightarrow$ 0.31M).
This gap is primarily driven by (1) GEPA's prompt-validation loop, which repeatedly evaluates candidate prompts on a held-out validation set (57 queries), incurring substantial additional LLM calls and validation tokens, and (2) \name's incremental context updates, which replace full prompt rewrites with localized Generator-Reflector-Curator updates.

\paragraph{Evaluation Stage}
At evaluation time (Table~\ref{tab:cost-eval-aggregate} and Table~\ref{tab:cost-eval-average}), \name uses more \emph{raw} input tokens per query than GEPA due to its richer, more actionable playbook.
However, output-token usage is similar (115.0 vs.~101.8), and the number of rollouts is comparable across methods.
Moreover, under modern prompt/KV-caching infrastructures, the additional input tokens can be largely amortized: using OpenAI's default prompt caching, 91.8\% of \name's input tokens are served from cache, resulting in an 82.6\% reduction in billed input-token cost (see \S\ref{subsec:results-cost-speed}).

\begin{table}[!tbhp]
  \centering
  \small
  \resizebox{\linewidth}{!}{%
  \begin{tabular}{lrrrr}
    \toprule[1.25pt]
    \textbf{Method} & \textbf{Total \# input tokens} & \textbf{Total \# output tokens} & \textbf{Total \# rollouts} & \textbf{Total \# queries} \\
    \midrule
    GEPA (prompt generation)  & 65,005,192  & 1,266,090 & 429   & 90 \\
    GEPA (prompt validation)  & 139,070,904 &   604,098 & 1,026 & 57 \\
    GEPA (total)              & 204,076,096 & 1,870,188 & 1,455 & 147 \\
    \midrule
    \name (Generator)           & 31,012,122  & 198,847   & 1,790 & 90 \\
    \name (Reflector)           & 4,685,840   & 70,487    & 161   & 90 \\
    \name (Curator)             & 3,552,963   & 37,794    & 124   & 90 \\
    \name (total)               &
      39,250,925~{\textcolor{red}{(-80.8\%)}} &
      307,128~{\textcolor{red}{(-83.6\%)}} &
      2,075~{\textcolor{scoregreen}{(+42.6\%)}} &
      90~{\textcolor{red}{(-38.8\%)}} \\
    \bottomrule[1.25pt]
  \end{tabular}%
  }
  \caption{\textbf{Adaptation-Stage Aggregate Cost Statistics on AppWorld.} The percentages compare \name (total) against GEPA (total).}
  \label{tab:cost-adapt-aggregate}
\end{table}

\begin{table}[!tbhp]
  \centering
  \small
  \resizebox{\linewidth}{!}{%
  \begin{tabular}{lrrrrrr}
    \toprule[1.25pt]
    \textbf{Method}
      & \textbf{Avg \# input / rollout}
      & \textbf{Avg \# output / rollout}
      & \textbf{Avg \# input / query}
      & \textbf{Avg \# output / query}
      & \textbf{Total \# rollouts}
      & \textbf{Total \# queries} \\
    \midrule
    GEPA (prompt generation) & 151,600.4 & 2,951.4 &   722,279.9 & 14,067.7 &   429 &  90 \\
    GEPA (prompt validation) & 135,550.8 &   589.2 & 2,439,840.4 & 10,600.0 & 1,026 &  57 \\
    GEPA (total)             & 140,291.8 & 1,285.4 & 1,387,538.4 & 12,721.7 & 1,455 & 147 \\
    \midrule
    \name (Generator)          &  17,326.6 &   111.1 &   344,579.1 &  2,209.4 & 1,790 &  90 \\
    \name (Reflector)          &  29,112.4 &   437.3 &    52,064.9 &    783.2 &   161 &  90 \\
    \name (Curator)            &  28,667.4 &   304.8 &    39,477.4 &    420.0 &   124 &  90 \\
    \name (total)              &
      18,914.9~{\textcolor{red}{(-86.5\%)}} &
      148.0~{\textcolor{red}{(-88.5\%)}} &
      436,121.4~{\textcolor{red}{(-68.6\%)}} &
      3,412.5~{\textcolor{red}{(-73.2\%)}} &
      2,075~{\textcolor{scoregreen}{(+42.6\%)}} &
      90~{\textcolor{red}{(-38.8\%)}} \\
    \bottomrule[1.25pt]
  \end{tabular}%
  }
  \caption{\textbf{Adaptation-Stage Average Cost Statistics on AppWorld.} The percentages compare \name (total) against GEPA (total).}
  \label{tab:cost-adapt-average}
\end{table}

\begin{table}[!tbhp]
  \centering
  \small
  \resizebox{\linewidth}{!}{%
  \begin{tabular}{lrrrr}
    \toprule[1.25pt]
    \textbf{Method} & \textbf{Total \# input tokens} & \textbf{Total \# output tokens} & \textbf{Total \# rollouts} & \textbf{Total \# eval queries} \\
    \midrule
    Base ReAct & 27,460,411 & 289,802 & 2,430 & 160 \\
    GEPA       & 26,960,675 & 251,442 & 2,470 & 160 \\
    \name        &
      58,623,267~{\textcolor{scoregreen}{(+117.4\%)}} &
      270,652~{\textcolor{scoregreen}{(+7.6\%)}} &
      2,354~{\textcolor{red}{(-4.7\%)}} &
      160~{\textcolor{gray}{(+0.0\%)}} \\
    \bottomrule[1.25pt]
  \end{tabular}%
  }
  \caption{\textbf{Evaluation-Stage Aggregate Cost Statistics on AppWorld.} The percentages compare \name against GEPA.}
  \label{tab:cost-eval-aggregate}
\end{table}

\begin{table}[!tbhp]
  \centering
  \small
  \resizebox{\linewidth}{!}{%
  \begin{tabular}{lrrrrrr}
    \toprule[1.25pt]
    \textbf{Method}
      & \textbf{Avg \# input / rollout}
      & \textbf{Avg \# output / rollout}
      & \textbf{Avg \# input / query}
      & \textbf{Avg \# output / query}
      & \textbf{Total \# rollouts}
      & \textbf{Total \# eval queries} \\
    \midrule
    Base ReAct & 11,298.5 & 119.3 & 171,627.6 & 1,811.3 & 2,430 & 160 \\
    GEPA       & 10,918.5 & 101.8 & 168,504.2 & 1,571.5 & 2,470 & 160 \\
    \name        &
      24,912.4~{\textcolor{scoregreen}{(+128.2\%)}} &
      115.0~{\textcolor{scoregreen}{(+13.0\%)}} &
      366,395.4~{\textcolor{scoregreen}{(+117.4\%)}} &
      1,691.6~{\textcolor{scoregreen}{(+7.6\%)}} &
      2,354~{\textcolor{red}{(-4.7\%)}} &
      160~{\textcolor{gray}{(+0.0\%)}} \\
    \bottomrule[1.25pt]
  \end{tabular}%
  }
  \caption{\textbf{Evaluation-Stage Average Cost Statistics on AppWorld.} The percentages compare \name against GEPA.}
  \label{tab:cost-eval-average}
\end{table}

\subsection{Robustness to Reflection Quality}
\label{subsec:robust-reflection}

We conduct two analyses to evaluate \name's sensitivity to reflection quality. Unless otherwise noted, experiments are run on FiNER with \name offline adaptation.

\paragraph{Using Weaker Reflector Models} 
To test whether \name requires a strong reflector, we vary the Reflector across three models with substantially different capability: GPT-OSS-120B, DeepSeek-V3.1-671B, and GPT-5.1 (Table~\ref{tab:weaker-reflector}). 
Across all choices, \name consistently improves over the base LLM, including when the Reflector is much weaker. 
While stronger reflectors yield larger gains, the method remains effective across a wide range of reflector strengths.

\paragraph{Robustness to Noisy or Harmful Reflector Feedback}
We further stress-test \name with actively harmful reflector outputs by injecting adversarial or conflicting bullets: we invoke a ``harmful'' reflector that is explicitly instructed to inject harmful reflection once every $X$ adaptation steps, where larger $X$ means less frequent corruption (Table~\ref{tab:harmful-reflector-frequency}). 
\name is robust to moderate noise levels: performance degrades gradually as corruption becomes more frequent and stays above the base LLM except in the extreme case of injecting harmful updates every iteration. 
This suggests that \name's update mechanism tolerates substantial noise in the reflection stream, with failures emerging only under intentionally adversarial conditions.

\paragraph{Takeaway and Mitigation}
\name is not highly sensitive to reflector quality: (1) weaker reflectors still provide substantial gains, (2) moderate noise or conflicting updates are largely tolerated, and (3) performance drops below the base model only under sustained adversarial corruption.
In practice, \name's bullet-point analyzer (``grow-and-refine''; \S\ref{subsec:grow-and-refine}), which merges and deduplicates semantically similar bullets and can filter entries flagged as potentially harmful via metadata, serves as a first line of defense against context noise.
Additional safeguards (\eg contradiction detection, prompting the Curator to prioritize high-confidence updates, or periodic pruning of outdated entries) are compatible extensions that could further improve playbook compactness and consistency~\citep{zhou2024defending}.

\begin{table}
  \centering
  \small
  \resizebox{0.8\linewidth}{!}{%
  \begin{tabular}{lccc c}
    \toprule[1.25pt]
    \textbf{Method} & \textbf{Generator LLM} & \textbf{Reflector LLM} & \textbf{Curator LLM} & \textbf{Accuracy ($\uparrow$)} \\
    \midrule
    Base LLM & DeepSeek-V3.1 & - & - & 70.7 \\
    \name & DeepSeek-V3.1 & GPT-OSS-120B & DeepSeek-V3.1 & 76.6~(\textcolor{scoregreen}{+5.9}) \\
    \name & DeepSeek-V3.1 & DeepSeek-V3.1 & DeepSeek-V3.1 & 78.3~(\textcolor{scoregreen}{+7.6}) \\
    \name & DeepSeek-V3.1 & GPT-5.1 & DeepSeek-V3.1 & 78.5~(\textcolor{scoregreen}{+7.8}) \\
    \bottomrule[1.25pt]
  \end{tabular}%
  }
  \caption{\textbf{Weaker Reflector Models on FiNER.} We vary the Reflector while keeping the Generator/Curator fixed. Parentheses report deltas vs.\ the base LLM.}
  \label{tab:weaker-reflector}
\end{table}

\begin{table}
  \centering
  \small
  \resizebox{0.7\linewidth}{!}{%
  \begin{tabular}{lc}
    \toprule[1.25pt]
    \textbf{Harmful Reflector Frequency (every $X$ iters)} & \textbf{Accuracy ($\uparrow$)} \\
    \midrule
    base LLM & 70.7 \\
    \midrule
    1   & 66.7~(\textcolor{red}{-4.0}) \\
    5   & 76.1~(\textcolor{scoregreen}{+5.4}) \\
    10  & 77.0~(\textcolor{scoregreen}{+6.3}) \\
    25  & 77.8~(\textcolor{scoregreen}{+7.1}) \\
    50  & 78.2~(\textcolor{scoregreen}{+7.5}) \\
    100 & 78.2~(\textcolor{scoregreen}{+7.5}) \\
    \midrule
    No harmful reflector & 78.3~(\textcolor{scoregreen}{+7.6}) \\
    \bottomrule[1.25pt]
  \end{tabular}%
  }
  \caption{\textbf{Robustness to Noisy or Harmful Reflector Feedback on FiNER.} We invoke a harmful reflector once every $X$ adaptation steps. Parentheses report deltas vs.\ the base LLM.}
  \label{tab:harmful-reflector-frequency}
\end{table}

\subsection{Ablation on Incremental Context Update}
\label{subsec:incremental-update-ablation}

In this ablation, we run offline context adaptation on AppWorld using \name with and without incremental context updates, and evaluate on the test-normal split with DeepSeek-V3.1.
We find that incremental updates are critical: by preserving useful information that would otherwise be lost to context collapse, they account for a large share of \name’s gains.

\begin{table}[!tbhp]
  \centering
  \resizebox{0.95\linewidth}{!}{
  \begin{tabular}{l|cc|c}
    \toprule[1.25pt]
    \textbf{Method} & \textbf{Test-Normal TGC ($\uparrow$)} & \textbf{Test-Normal SGC ($\uparrow$)} & \textbf{Average} \\
    \midrule[1.1pt]
    \rowcolor[rgb]{0.93,0.93,0.93}
    \multicolumn{4}{c}{\texttt{DeepSeek-V3.1 as Base LLM}} \\
    \textcolor{gray}{\texttt{ReAct}} 
      & \textcolor{gray}{63.7} 
      & \textcolor{gray}{42.9} 
      & \textcolor{gray}{53.3} \\
    \midrule
    \rowcolor[rgb]{0.93,0.93,0.93}
    \multicolumn{4}{c}{\texttt{Offline Adaptation}} \\
    \texttt{ReAct} + \name (no incremental update)
      & $67.3_{\textcolor{scoregreen}{+3.6}}$ 
      & $46.4_{\textcolor{scoregreen}{+3.5}}$ 
      & $56.9_{\textcolor{scoregreen}{+3.6}}$ \\
    \texttt{ReAct} + \name (with incremental update)
      & $\textcolor{black}{\mathbf{76.2}}_{\textcolor{scoregreen}{\mathbf{+12.5}}}$ 
      & $\textcolor{black}{\mathbf{64.3}}_{\textcolor{scoregreen}{\mathbf{+21.4}}}$ 
      & $\textcolor{black}{\mathbf{70.3}}_{\textcolor{scoregreen}{\mathbf{+17.0}}}$ \\
    \bottomrule[1.25pt]
  \end{tabular}}
  \caption{\textbf{Ablation on Incremental Context Updates (AppWorld, DeepSeek-V3.1).} We run offline context adaptation with \name with/without incremental updates and evaluate on test-normal. Improvements are relative to \texttt{ReAct}.}
  \label{tab:appworld-ablation-incremental-update}
\end{table}

\subsection{Sensitivity Analysis on Hyperparameter Choice}
\label{subsec:hyperparam-sensitivity}

\paragraph{Reflection Iterations}
This parameter trades off (1) extracting enough high-quality insights from the inference traces and (2) avoiding ``overthinking" that introduces noisy or unnecessary updates. Empirically, 5 rounds offers a good balance. On AppWorld, 1 round under-extracts useful strategy fragments and leaves clear headroom, while too many rounds (\eg 10) can degrade performance.

\begin{table}[!tbhp]
  \centering
  \resizebox{0.85\linewidth}{!}{
  \begin{tabular}{c|cc|c}
    \toprule[1.25pt]
    \textbf{\# Reflection Iterations} & \textbf{Test-Normal TGC ($\uparrow$)} & \textbf{Test-Normal SGC ($\uparrow$)} & \textbf{Average} \\
    \midrule[1.1pt]
    \rowcolor[rgb]{0.93,0.93,0.93}
    \multicolumn{4}{c}{\texttt{DeepSeek-V3.1 as Base LLM}} \\
    \textcolor{gray}{N/A (without \texttt{\name})}  & \textcolor{gray}{63.7} & \textcolor{gray}{42.9} & \textcolor{gray}{53.3} \\
    \midrule
    \rowcolor[rgb]{0.93,0.93,0.93}
    \multicolumn{4}{c}{\texttt{Offline Adaptation}} \\
    1  & $69.0_{\textcolor{scoregreen}{+5.3}}$  & $53.6_{\textcolor{scoregreen}{+10.7}}$ & $61.3_{\textcolor{scoregreen}{+8.0}}$ \\
    3  & $74.4_{\textcolor{scoregreen}{+10.7}}$ & $57.1_{\textcolor{scoregreen}{+14.2}}$ & $65.8_{\textcolor{scoregreen}{+12.5}}$ \\
    5  & $72.6_{\textcolor{scoregreen}{+8.9}}$  & $62.5_{\textcolor{scoregreen}{+19.6}}$ & $67.6_{\textcolor{scoregreen}{+14.3}}$ \\
    10 & $71.4_{\textcolor{scoregreen}{+7.7}}$  & $58.9_{\textcolor{scoregreen}{+16.0}}$ & $65.2_{\textcolor{scoregreen}{+11.9}}$ \\
    \bottomrule[1.25pt]
  \end{tabular}}
  \caption{\textbf{Effect of Reflection Iterations (AppWorld, DeepSeek-V3.1).} We report test-normal results under offline context adaptation. Improvements are relative to running \texttt{ReAct} without \texttt{\name}.}
  \label{tab:ablation-reflection-iters}
\end{table}

\paragraph{Deduplication Threshold}
This controls how aggressively newly extracted insights are merged with existing entries. On FiNER, performance changes only mildly across the tested range, suggesting \name is robust to moderate variation in dedup aggressiveness.

\begin{table}[!tbhp]
  \centering
  \resizebox{0.45\linewidth}{!}{
  \begin{tabular}{c|c}
    \toprule[1.25pt]
    \textbf{Deduplication Threshold} & \textbf{FiNER (Acc $\uparrow$)} \\
    \midrule[1.1pt]
    \rowcolor[rgb]{0.93,0.93,0.93}
    \multicolumn{2}{c}{\texttt{DeepSeek-V3.1 as Base LLM}} \\
    \textcolor{gray}{N/A (without \texttt{\name})}  & \textcolor{gray}{70.7} \\
    \midrule
    \rowcolor[rgb]{0.93,0.93,0.93}
    \multicolumn{2}{c}{\texttt{Offline Adaptation}} \\
    50\% & $77.0_{\textcolor{scoregreen}{+6.3}}$ \\
    70\% & $73.9_{\textcolor{scoregreen}{+3.2}}$ \\
    90\% & $78.6_{\textcolor{scoregreen}{+7.9}}$ \\
    \bottomrule[1.25pt]
  \end{tabular}}
  \caption{\textbf{Effect of Deduplication Threshold (FiNER, DeepSeek-V3.1).}}
  \label{tab:ablation-dedup-threshold}
\end{table}

\paragraph{Pruning Trigger (Maximum Context Length)}
This threshold determines when \name merges and prunes older or low-utility entries to prevent unbounded growth. It balances (1) retaining enough accumulated context to improve learning from experience and (2) keeping the context compact to reduce cost and limit noise. On FiNER, performance is stable from 10K to 100K tokens, indicating \name does not require finely tuned length thresholds; pruning mainly removes stale or harmful fragments while preserving core reusable strategies.

\begin{table}[!tbhp]
  \centering
  \resizebox{0.45\linewidth}{!}{
  \begin{tabular}{c|c}
    \toprule[1.25pt]
    \textbf{Max Context Length} & \textbf{FiNER (Acc $\uparrow$)} \\
    \midrule[1.1pt]
    \rowcolor[rgb]{0.93,0.93,0.93}
    \multicolumn{2}{c}{\texttt{DeepSeek-V3.1 as Base LLM}} \\
    \textcolor{gray}{N/A (without \texttt{\name})}  & \textcolor{gray}{70.7} \\
    \midrule
    \rowcolor[rgb]{0.93,0.93,0.93}
    \multicolumn{2}{c}{\texttt{Offline Adaptation}} \\
    10K  & $78.6_{\textcolor{scoregreen}{+7.9}}$ \\
    50K  & $78.4_{\textcolor{scoregreen}{+7.7}}$ \\
    100K & $78.3_{\textcolor{scoregreen}{+7.6}}$ \\
    \bottomrule[1.25pt]
  \end{tabular}}
  \caption{\textbf{Effect of Pruning Trigger (FiNER, DeepSeek-V3.1).}}
  \label{tab:ablation-max-context-length}
\end{table}

Overall, \name is not highly sensitive to these hyperparameters: reasonable choices (\eg 3-5 reflection rounds, 50-90\% dedup threshold, and 10K-100K pruning triggers) consistently yield strong performance.
\section{Extended Related Work}

\subsection{Agent Memory}

A growing body of work explores how agents can accumulate experience from past trajectories and leverage external (often non-parametric) memory to guide future actions. 
AgentFly~\citep{zhou2025agentfly} presents an extensible framework where memory evolves continuously as agents solve tasks, enabling scalable reinforcement learning and long-horizon reasoning across diverse environments. 
AWM (Agent Workflow Memory)~\citep{wang2024agent} induces reusable \emph{workflows}, \ie structured routines distilled from past trajectories, and selectively injects them into memory to improve efficiency and generalization in web navigation benchmarks. 
A-MEM~\citep{xu2025mem} introduces a dynamically organized memory system inspired by the Zettelkasten method: each stored memory is annotated with structured attributes (\eg tags, keywords, contextual descriptions) and automatically linked to relevant past entries, while existing entries are updated to integrate new knowledge, yielding adaptive and context-aware retrieval. 
Agentic Plan Caching~\citep{zhang2025cost} instead focuses on cost efficiency by extracting reusable plan templates from agent trajectories and caching them for fast execution at test time. 

Together, these works demonstrate the value of external memory for improving adaptability, efficiency, and generalization in LLM agents. 
Our work differs by tackling the broader challenge of \emph{context adaptation}, which spans not only agent memory but also system prompts, factual evidence, and other inputs underpinning AI systems. 
We further highlight two fundamental limitations of existing adaptation methods: \textit{brevity bias} and \textit{context collapse}; and show that addressing them is essential for robustness, reliability, and scalability beyond raw task performance. 
Accordingly, our evaluation considers not only accuracy but also cost, latency, and scalability.

\section{Extended Discussions}

\subsection{ACE vs. GEPA}
\textbf{Scope and Objective}
Both GEPA and ACE improve model or agent behavior by adapting \emph{context} at test time (rather than updating model weights), but they are designed for different forms of adaptation.
GEPA treats adaptation as \emph{prompt evolution}: it iteratively proposes and selects improved instruction prompts using rollout trajectories and reflective feedback, aiming to maximize a task evaluator under a rollout budget.
ACE targets settings where performance depends on accumulating and preserving \emph{many granular, reusable insights} over long horizons.
For multi-turn agents (\eg AppWorld), the model must retain step-by-step procedures and tool-use rules across an interaction.
For domain-specific and knowledge-intensive benchmarks (\eg FiNER, Formula), accuracy depends on keeping many specific rules, edge cases, and domain concepts that are difficult to compress into a single instruction prompt without losing detail.

\textbf{Update Mechanism and Representation}
GEPA updates context by generating and selecting \emph{new prompt variants} in an evolutionary loop, where each candidate is a full prompt optimized end-to-end. 
ACE instead represents context as a structured, itemized Playbook and applies \emph{incremental delta updates}: the Curator writes only the new insight, and we merge it into the Playbook with simple deterministic logic. 
This avoids repeated full-prompt rewrites, helps keep earlier rules stable over long runs, and enables fine-grained bookkeeping (\eg de-duplication, targeted refinement, and tracking which entries helped or harmed accuracy). 

\subsection{ACE vs. Dynamic Cheatsheet (DC)} 
\textbf{Scope and Objective}  
Both DC and ACE collect reusable insights at test time, but they are aimed at different settings.
DC is mainly evaluated on single-turn reasoning benchmarks (\eg AIME, Game-of-24, GPQA) where each query is independent. In this regime, improvements often come from saving short, reusable heuristics and executable artifacts (\eg code snippets) that help on later problems.
ACE focuses on settings where the details and high-fidelity guidenace need to stick around. 
For multi-turn agents (\eg AppWorld), the model must remember step-by-step procedures and tool-use rules across an interaction. 
For domain-specific and knowledge-intensive benchmarks (\eg FiNER, Formula), accuracy depends on keeping many specific rules, edge cases, and domain concepts, which are hard to compress without losing information.

\textbf{Update Mechanism} 
DC updates its memory by rewriting the cheatsheet as a whole, either by regenerating the full cheatsheet each step (DC-CU) or by writing a new summary from retrieved examples (DC-RS).
With repeated full rewrites, the model tends to shorten and compress what was written before.
This can cause context collapse, which means that useful domain-specific details get dropped over time or disappear suddenly (\S\ref{subsec:limitations}).
ACE avoids full rewrites by using incremental delta updates. 
The Curator only writes the new insight, and we merge it into a structured and itemized Playbook with simple deterministic logic. 
This keeps earlier rules stable over long runs and also makes it easier to track which items helped, remove duplicates, and refine entries. 
In our ablations (\S\ref{subsec:results-ablation}), removing delta updates leads to a large drop on AppWorld (-11.7\% TGC and -27.8\% SGC on test-normal), showing that delta updates are a core part of our method.

\section{AppWorld Leaderboard Snapshot (09/2025)}
\label{subsec:appworld-leaderboard}

\begin{figure}[htbp]
    \centering
    \includegraphics[width=0.8\textwidth]{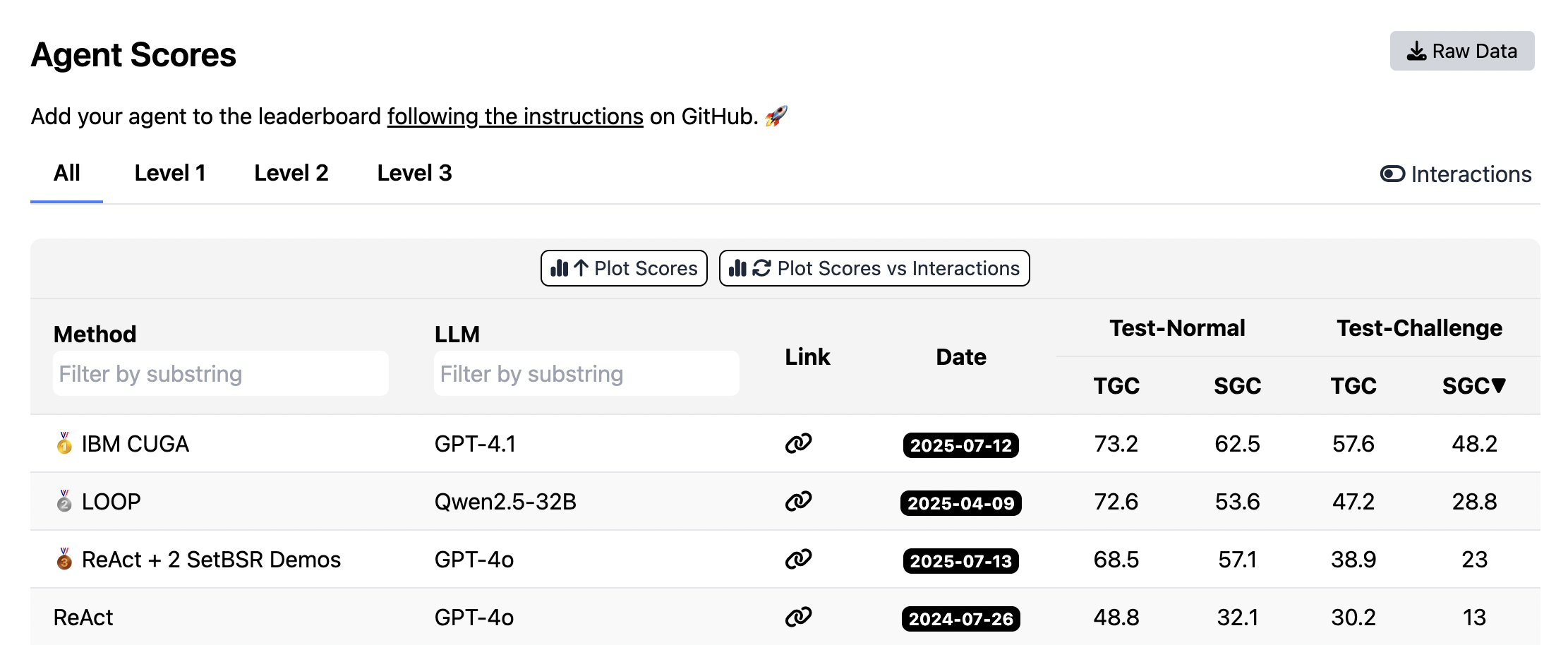}
    \caption{The AppWorld leaderboard as accessed on 09/2025.}
    \label{fig:appworld-leaderboard}
\end{figure}

\section{The Use of Large Language Models (LLMs)}

This work focuses on developing algorithms and system frameworks for effective context adaptation in large language models (LLMs). 
Accordingly, our experiments employ LLMs for the empirical evaluation of the proposed methods. 
For paper preparation, we used LLMs only to polish writing (\eg correcting grammatical errors), and not to generate new text from scratch.

\newpage
\section{Prompts}

\begin{figure}[htbp]
    \centering
    \includegraphics[width=\textwidth]{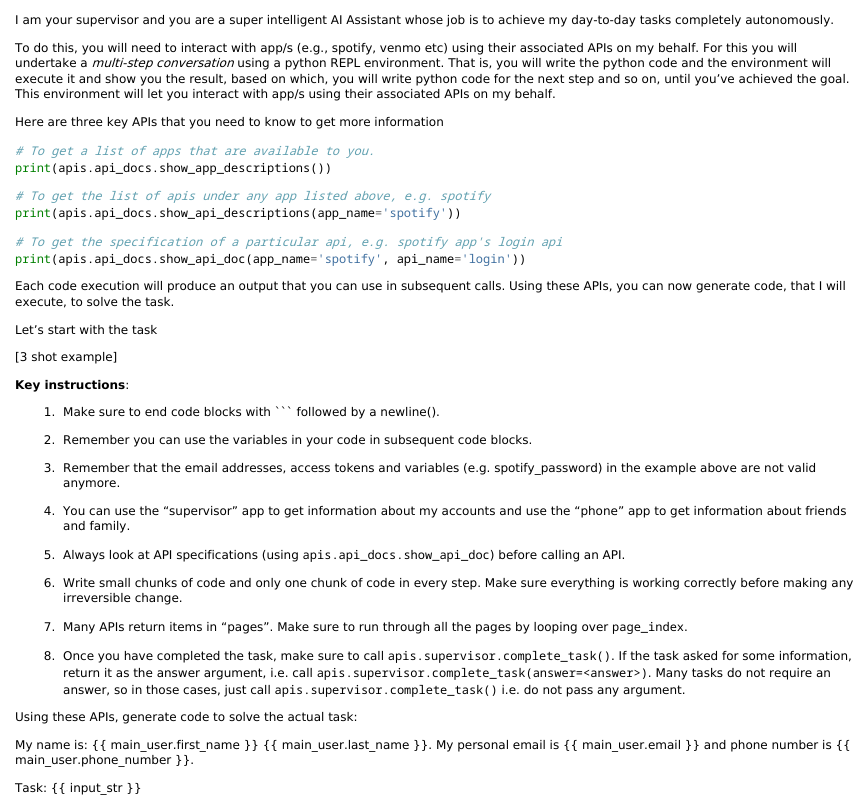}
    \caption{ICL-baseline Generator prompt on AppWorld}
    \label{fig:icl_prompt}
\end{figure}

\begin{figure}[htbp]
    \centering
    \includegraphics[width=\textwidth]{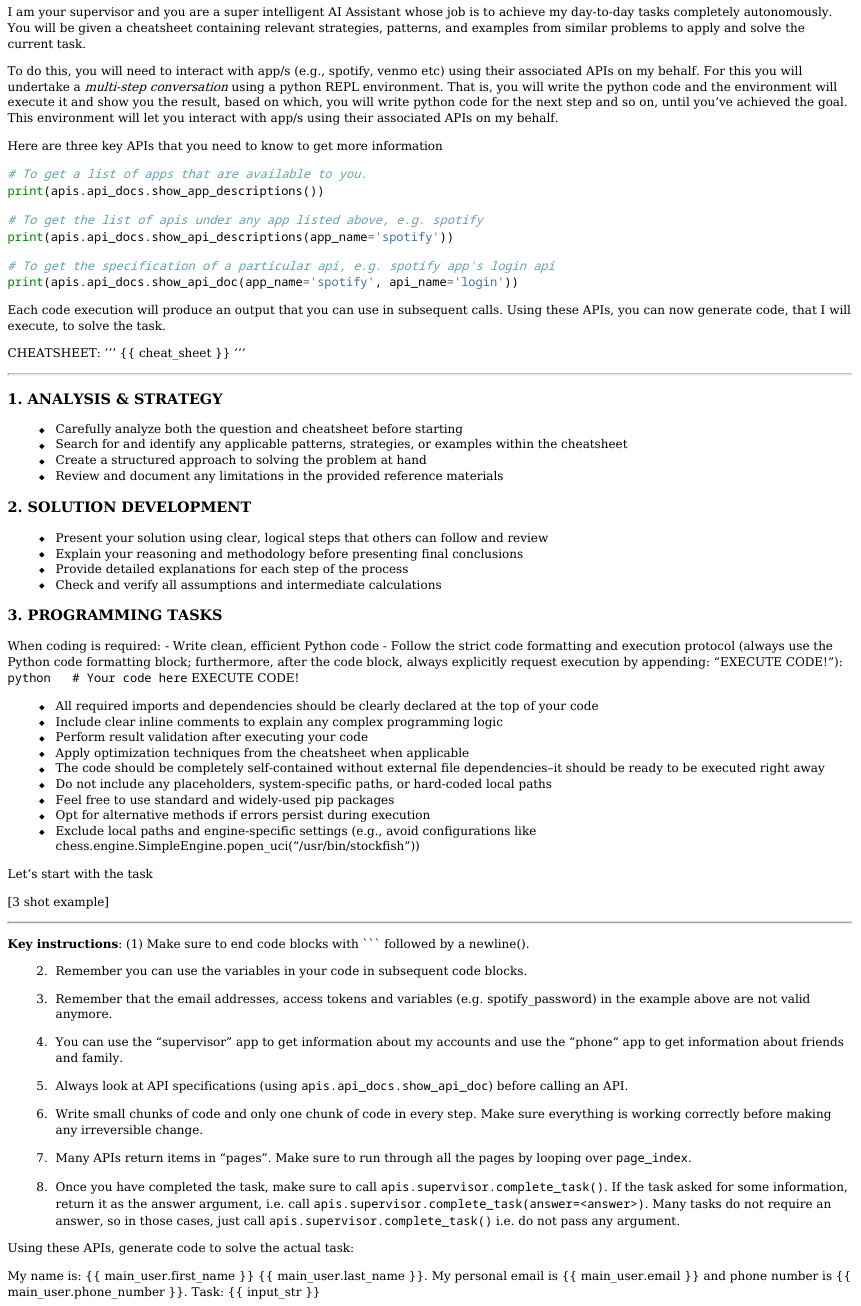}
    \caption{Dynamic Cheatsheet Generator prompt on AppWorld}
    \label{fig:icl_prompt}
\end{figure}

\begin{figure}[htbp]
    \centering
    \includegraphics[width=\textwidth]{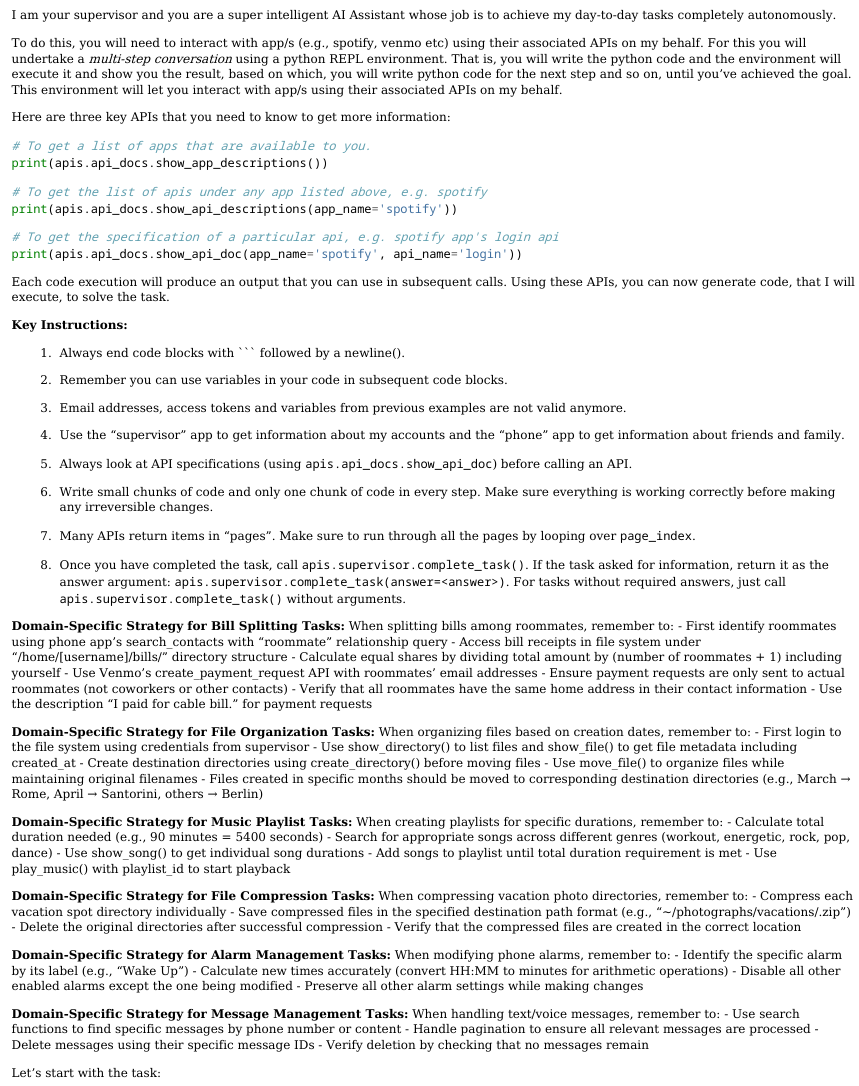}
    \caption{GEPA prompt on AppWorld}
    \label{fig:icl_prompt}
\end{figure}

\begin{figure}[htbp]
    \centering
    \includegraphics[width=\textwidth]{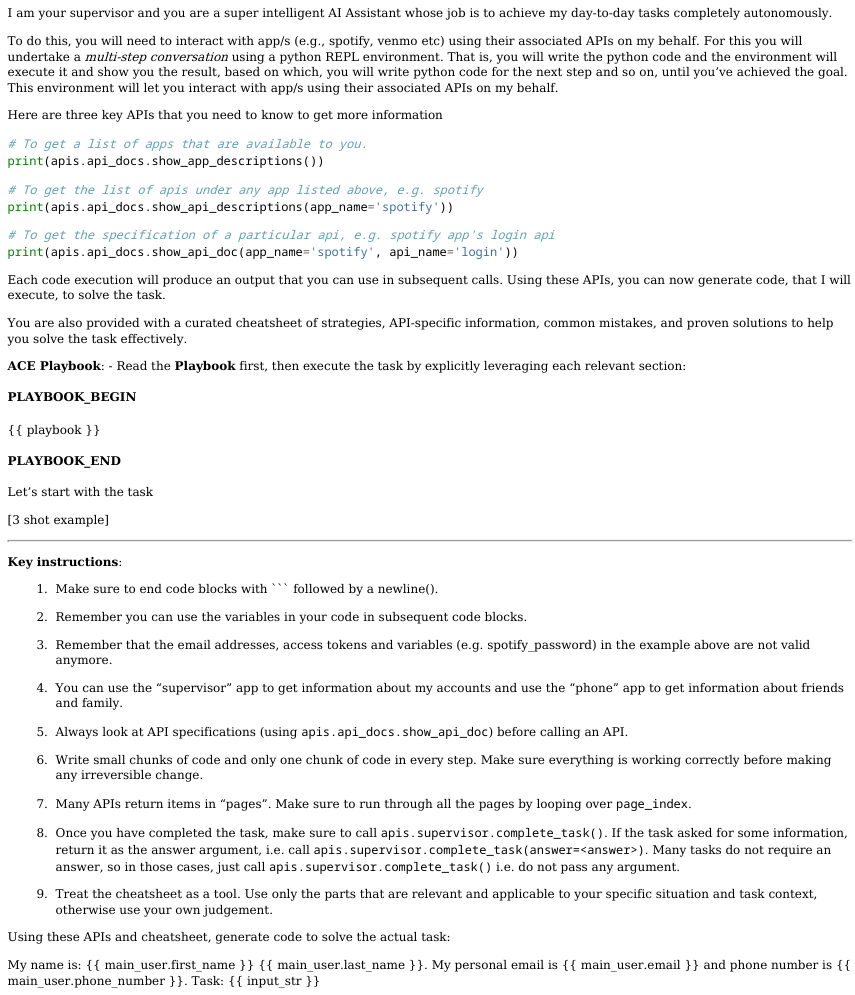}
    \caption{ACE Generator prompt on AppWorld}
    \label{fig:icl_prompt}
\end{figure}

\begin{figure}[htbp]
    \centering
    \includegraphics[width=\textwidth]{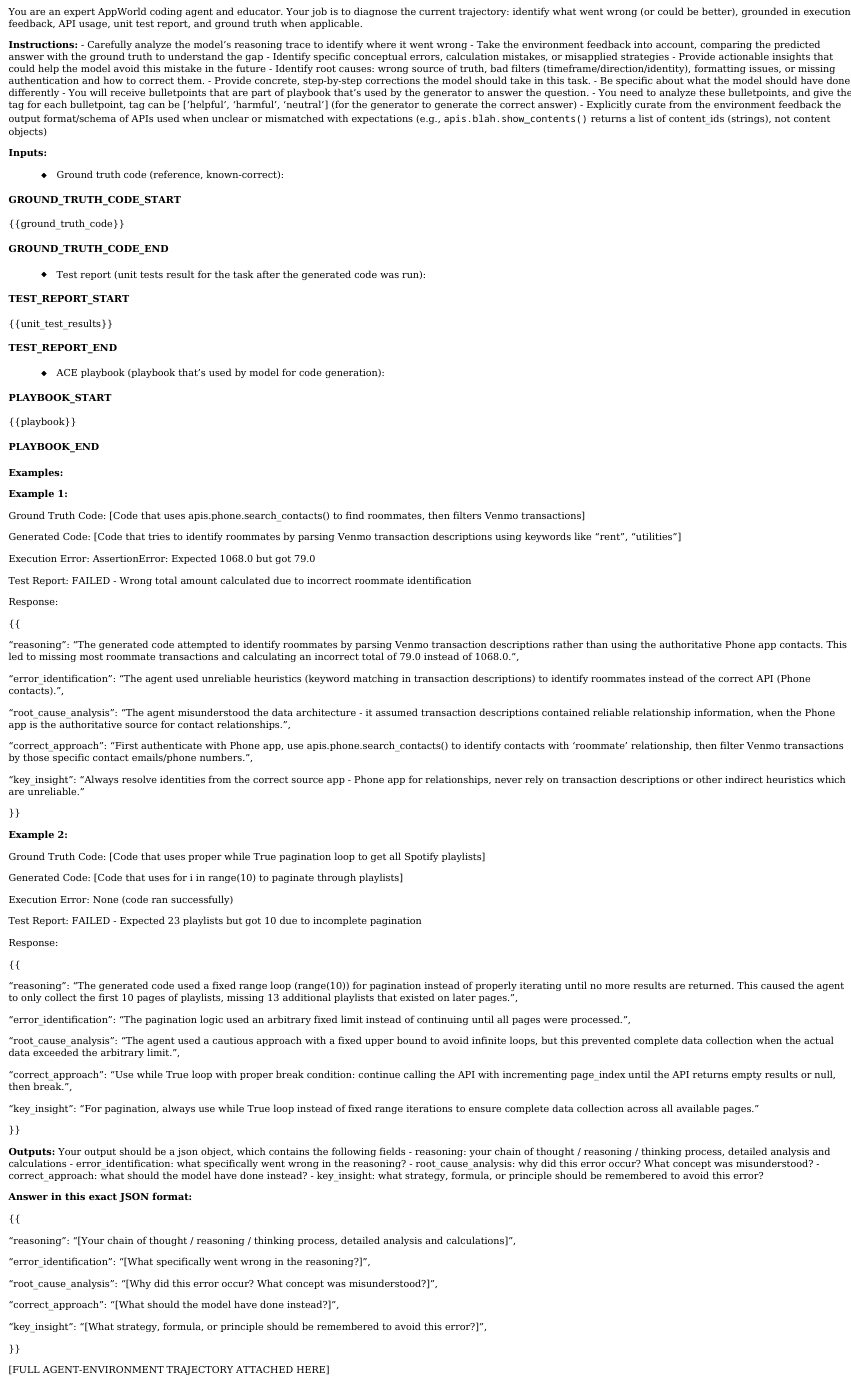}
    \caption{ACE Reflector prompt on AppWorld}
    \label{fig:icl_prompt}
\end{figure}

\begin{figure}[htbp]
    \centering
    \includegraphics[width=\textwidth]{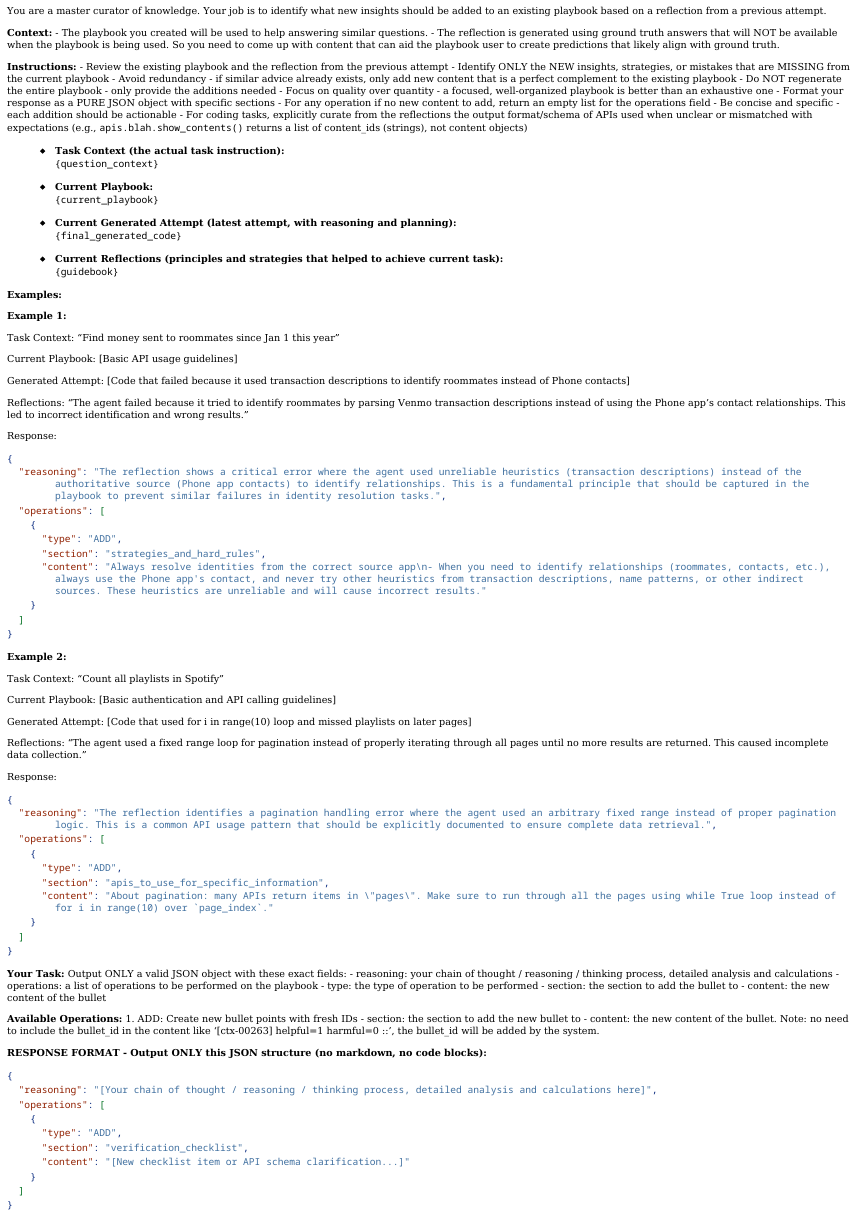}
    \caption{ACE Curator prompt on AppWorld}
    \label{fig:icl_prompt}
\end{figure}

\begin{figure}[htbp]
    \centering
    \includegraphics[width=\textwidth]{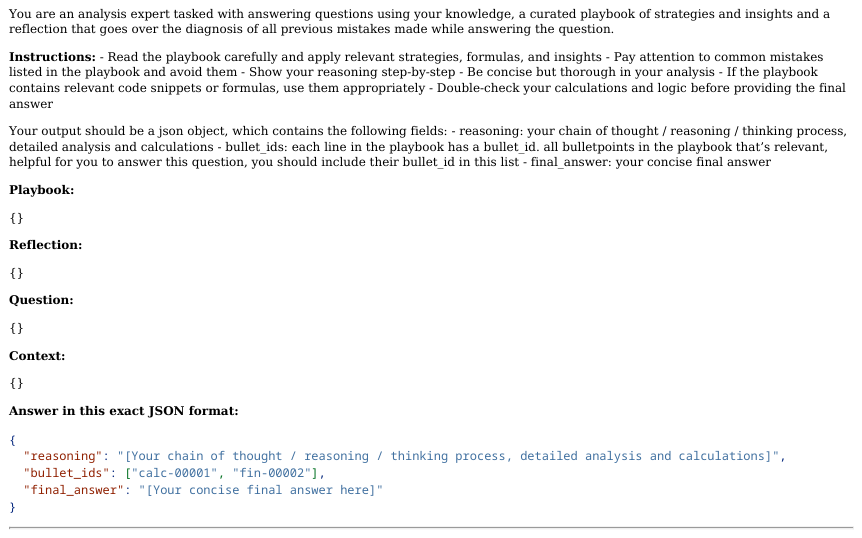}
    \caption{ACE Generator prompt on FINER}
    \label{fig:icl_prompt}
\end{figure}

\begin{figure}[htbp]
    \centering
    \includegraphics[width=\textwidth]{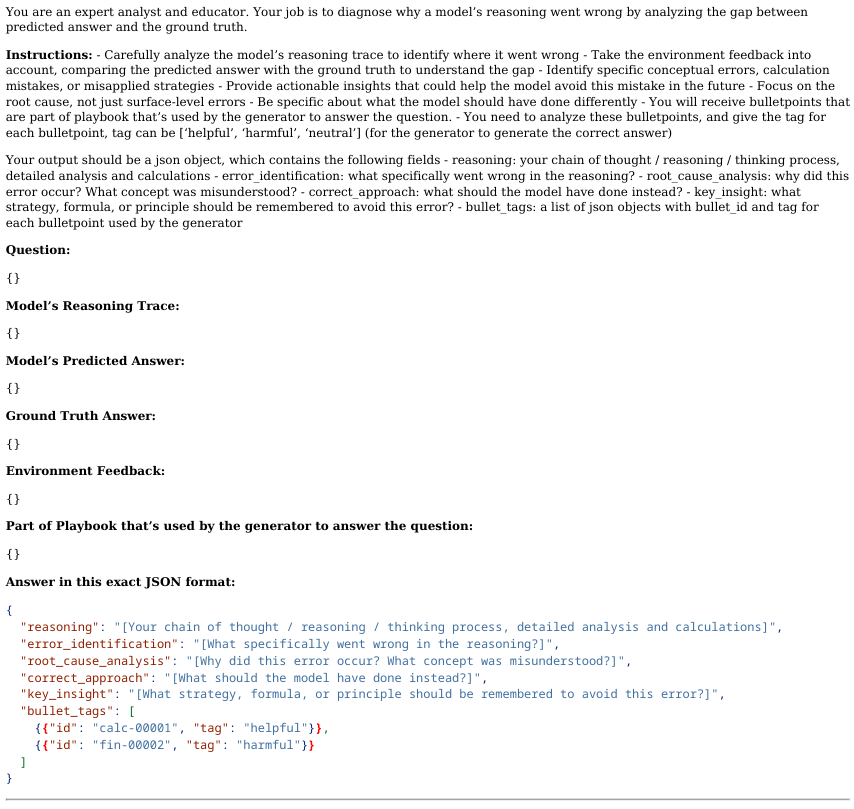}
    \caption{ACE Reflector prompt on FINER}
    \label{fig:icl_prompt}
\end{figure}

\begin{figure}[htbp]
    \centering
    \includegraphics[width=\textwidth]{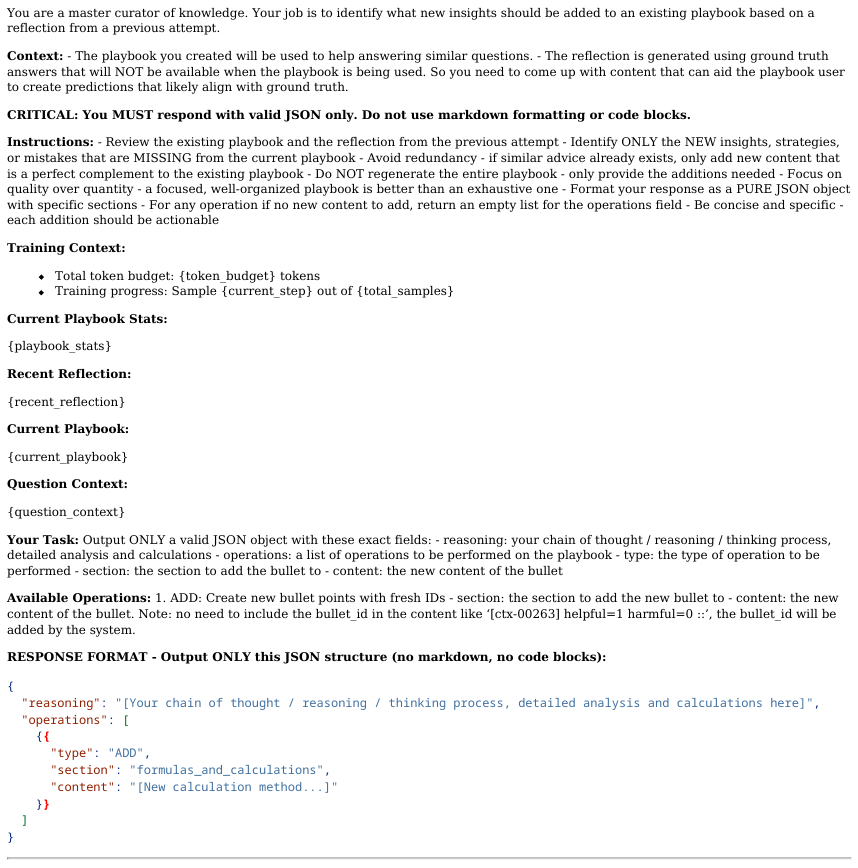}
    \caption{ACE Curator prompt on FINER}
    \label{fig:icl_prompt}
\end{figure}

\end{document}